\documentclass[12pt]{article}
\usepackage{authblk}
\usepackage{amsmath}
\usepackage{mathtools}
\mathtoolsset{showonlyrefs}
\usepackage{natbib}
\setcitestyle{authoryear}
\usepackage[geometry]{ifsym}
\usepackage{tikz-cd}
\usepackage{amsfonts}
\usepackage[margin=1in,showframe=false]{geometry}
\usepackage[hidelinks]{hyperref}

\def\R{{\mathbb R}}
\def\F{{\mathcal F}}
\def\N{{\mathcal N}}
\def\E{{\mathcal E}}
\def\L{{\mathcal L}}
\def\u{{\mathbf u}}
\def\cc{{\mathbf c}}
\def\z{{\mathbf z}}
\def\d{{\mathbf d}}
\def\w{{\mathbf w}}
\def\U{{\mathbf U}}
\def\bfxi{{\mathbf{\xi}}}
\DeclareMathOperator*{\mn}{min}

\begin{document}

\title{Solving High-Dimensional Inverse Problems with Auxiliary Uncertainty via Operator Learning with Limited Data}
\author[1]{J. Hart}
\author[2]{M. Gulian}
\author[1]{I. Manickam}
\author[1]{L. Swiler}
\affil[1]{\footnotesize Sandia National Laboratories, P.O. Box 5800, Albuquerque, NM 87123-1320}
\affil[2]{\footnotesize Sandia National Laboratories, PO Box 969, Livermore, CA 94551-0969}
\affil[ ]{\texttt{\{joshart,mgulian,imanick,lpswile\}@sandia.gov}}

\maketitle

\abstract{In complex large-scale systems such as climate, important effects are caused by a combination of confounding processes that are not fully observable. The identification of sources from observations of system state is vital for attribution and prediction, which inform critical policy decisions.  The difficulty of these types of inverse problems lies in the inability to isolate sources and the cost of simulating computational models. Surrogate models may enable the many-query algorithms required for source identification, but data challenges arise from high dimensionality of the state and source, limited ensembles of costly model simulations to train a surrogate model, and few and potentially noisy state observations for inversion due to measurement limitations. The influence of auxiliary processes adds an additional layer of uncertainty that further confounds source identification. We introduce a framework based on (1) calibrating deep neural network surrogates to the flow maps provided by an ensemble of simulations obtained by varying sources, and (2) using these surrogates in a Bayesian framework to identify sources from observations via optimization. Focusing on an atmospheric dispersion exemplar, we find that the expressive and computationally efficient nature of the deep neural network operator surrogates in appropriately reduced dimension allows for source identification with uncertainty quantification using limited data. Introducing a variable wind field as an auxiliary process, we find that a Bayesian approximation error approach is essential for reliable source inversion when uncertainty due to wind stresses the algorithm.}

\section{Introduction}\label{sec:intro}
Computational models for large-scale geophysical systems such as the atmosphere, oceans, and ice sheets have ushered in a new era of science which is critical to answer humanity's most pressing questions. 
Characterization of indeterminate or partially observed processes that influence the states of such systems plays a pivotal role informing major policy decisions~\citep{ipcc_2021_policy}. 
Such processes include, e.g., aerosol injection and interaction in the atmosphere, ocean carbon sequestration, and ice sheet basal mechanics. 
Within earth system models (ESMs), which simulate the coupled physical, chemical, and biological processes that drive the environment of the earth, the evolution and impact of these processes depend on large numbers of parameters~\citep{doe_earth_systems_model,heavens_2013}. 
However, for all of the advances in ESMs, their use is still limited by their computational complexity. Thus, the inverse problem to estimate climate processes from observed data remains a significant challenge because the model evaluations required to calibrate their parameterizations are limited in number.

Recent advances in scientific machine learning have paved the way to approximate states in complex systems, such as in climate and earth system models, using fast-to-evaluate reduced order models for various governing operators  \citep{asch2022toolbox}. In this article, we build on recent research in deep neural network (DNN) operator approximations by utilizing such approximations to enable inverse problems that would otherwise be computationally intractable if constrained by a ESM. The concept of mitigating computational cost in many-query analysis via surrogate modeling is not new; rather, it has been the focus of extensive research in the uncertainty quantification community~\citep{uq_handbook_surrogates}. The novelty in this work is a shifting of focus from traditional surrogate models for a low-dimensional set of quantities of interest to reduced order models for operators defined in the high-dimensional state space of the physical process. This shift enables utilization of spatiotemporal data which is essential to accurately reconstruct unobserved spatiotemporal processes. By combining recent advances in DNN operator approximation with Bayesian inversion, we obtain a pragmatic approach to facilitate inverse problems for large-scale geophysical systems. 

Our use of surrogate operators in state space falls within the burgeoning field of operator learning. To learn a map between functions, it is necessary to replace the infinite-dimensional function spaces representing the domain and range of the target operator by finite-dimensional spaces. Operator learning methods based on DNN approximation of the corresponding operator between the finite-dimensional spaces have shown promise in representing complex, nonlinear operators in dimensions that prohibit the use of classical parametrizations. These methods feature different discretization and dimension reduction techniques as well as architectures and training schemes for the DNN. They include modal methods \citep{patel2018nonlinear,patel2021physics,qin2019data,li2020neural}, graph based methods \citep{trask2022enforcing, gao2020physics, shukla2022scalable}, principal component analysis based methods \citep{bhattacharya2020model, hesthaven_2018}, meshless methods \citep{trask2019gmlsnets}, trunk-branch based methods \citep{lu2019deeponet,cai2021deepm}, and time-stepping methods \citep{You2021,Long2018,qin2019data}. Generally, they make use of ensembles of input-output function pairs to train the operator surrogate. They can be purely data-driven, though incorporation of physical knowledge has been studied by \citet{wang2021learning}. The proposed algorithms can be utilized to obtain operator surrogates for models that are fully specified but expensive, or for partial differential equation (PDE) model discovery \citep{patel2021physics} when certain parameters or functional forms are not known. They also differ in the type of noise in the data that can be admitted in each method; see \citet{patel2022error} for a detailed discussion. 

Examples of inverse problems across the geosciences include the estimation of basal dynamics of the Antarctic ice sheet~\citep{isaac2015scalable}, the determination of water contaminants in subsurface flow~\citep{laird_2005}, mechanics of plate coupling in mantle flow~\citep{ratnaswamy_2015}, estimation of subsurface material properties via full waveform inversion~\citep{fwi_2014}, and inversion in atmospheric transport~\citep{enting2002inverse}. Solving such inverse problems requires many model evaluations to search over the parameter space \citep{tarantola2005inverse}. When inverting for very high-dimensional parameterizations, such as discretized functions which vary in time and/or space, brute force approaches are prohibitively expensive. Optimization and derivative-based sampling algorithms have seen great advances to enable the efficient solution of inverse problems~\citep{ghattas_infinite_dim_bayes_1,ghattas_infinite_dim_bayes_2,cui_2014,Biegler_11} by leveraging derivative information to guide exploration of the parameter space. Nonetheless, for many problems involving climate and earth systems, the model complexity and sophistication of the solvers pose barriers, requiring intrusive and specialized software development to obtain derivative information needed for constrained optimization in these approaches. We seek to mitigate these issues using learned operators, leveraging algorithmic differentiation capabilities to enable efficient optimization and sampling in a non-intrusive method that is transferable to different applications.

The important characteristics which define the inverse problems under our consideration are:
\begin{enumerate}
\item Large-scale spatiotemporal processes.
\item Presence of multiple parameters and/or processes being sources of uncertainty.
\item Computational cost limiting the number of high-fidelity simulations.
\item Inability to intrude in the simulation code to extract derivative information.
\item Uncertainty quantification being crucial due to high-dimensions and data limitations.
\item Requirement to reduce parameter tuning due to the complexity of composing algorithms.
\end{enumerate}
With this backdrop of challenges, we propose to use operator learning to build efficient surrogate models that capture spatiotemporal dynamics from limited data. With the learned operator, Bayesian inversion is used to estimate parameters and quantify their uncertainty. Derivative-based optimization and sampling algorithms leverage algorithmic differentiation tools from the learned operator to efficiently generate approximate samples from the Bayesian posterior with minimal parameter tuning.

\section{Problem Formulation and Solution Outline} \label{sec:formulation}
We consider time-dependent models of the form
\begin{align} \label{eqn:dyn_pde}
& \frac{\partial u}{\partial t} + \mathcal A(u,z,w) = f(z,w) & \qquad \text{on } \Omega \times (0,\infty) \\
& \mathcal B(u,z,w) = 0 & \text{on } \partial \Omega \times (0,\infty) \\
& u= u_0 & \text{on } \Omega \times \{0\}
\end{align}
where $u:\Omega \times [0,\infty) \to \R$ represents a state which is defined for time $t \ge 0$ on a spatial domain $\Omega \subset \R^d$, with $d=1,2$, or $3$. The dynamics of $u$ depend on the differential operator $\mathcal A$, which may be linear or nonlinear, the boundary condition $\mathcal B$, the initial condition $u_0$, and the forcing term $f$. They also depend on $z$ and $w$ which may be scalars, vectors, or functions defined in time and/or space. We focus on the latter case. We assume that a final time $T>0$ has been specified as our focus is on solving inverse problems for which a finite time horizon of data is available. 

The parameter $z$ is the primary quantity of interest. We seek to estimate it by solving an inverse problem which combines the computational model~\eqref{eqn:dyn_pde} with sparse and noisy observations of the state variable $u$. The other parameter $w$, sometimes referred to as a nuisance parameter, is also assumed to be uncertain but is not of primary interest. Accommodating for uncertainty in $w$ when estimating $z$ is crucial, as will become evident throughout the article. 
We assume that the computational model~\eqref{eqn:dyn_pde} may be solved for any $z$ and $w$, and that we have access to an ensemble of such simulations. However, in many applications, the number of simulations may be limited by the computational complexity of the model. In such cases, attacking the inverse problem using brute force, by solving~\eqref{eqn:dyn_pde} for an array of parameters $z$ and $w$ in order to select the parameters which are most consistent with the observed data, is computationally infeasible. 

To overcome this challenge, we utilize a DNN approximation of the flow map of the model~\eqref{eqn:dyn_pde}, i.e., the operator mapping the state and parameters at time $t$ to the state at time $t+\Delta t$. We can quickly evaluate and compose the flow map with itself to obtain a surrogate operator for the evolution operator, enabling efficient model queries of the state. The architecture of the surrogate flow map, which includes dimension reduction and reconstruction of the state, as well as the algorithm to train it from an ensemble of simulations of the model ~\eqref{eqn:dyn_pde} are described in Section \ref{sec:flow_map_learning}. 
To solve the inverse problem, in Section \ref{sec:bayes_inv}, we use the surrogate flow map to facilitate efficient inversion within a Bayesian framework by leveraging automatic differentiation algorithms for optimization and sampling. 

In our numerical example presented in Section~\ref{sec:numerics}, the state $u$ corresponds to the concentration of $\text{SO}_2$ in an atmospheric dispersion model, $z$ to a partially known $\text{SO}_2$ source, and $w$ to atmospheric winds. We note that a more general formulation with multiple states $u:\Omega \times [0,T] \to \R^k$ for $k>1$ is possible, but we consider the scalar case for simplicity in this article. Our developments are motivated by a goal to solve source inversion problems using earth system models. This is discussed further in the conclusion, Section \ref{sec:conclusion}. 

From here forward, we will work with a discretization of the PDE model \eqref{eqn:dyn_pde}. Let 
\begin{equation}\label{eq:time_notes}
t_n, \quad n=0,1,2,\dots,N
\end{equation}
denote increasing points in time, with $t_0=0$ and $t_N=T$. We assume that we have access to an ensemble of parameters and corresponding states at these time intervals, and seek to reconstruct a source from observations at the same intervals. In a numerical method for solving \eqref{eqn:dyn_pde}, these nodes are placed in $[0,T]$ with a sufficiently large $N$ to ensure accuracy of the discretization. However, we do not constrain the nodes \eqref{eq:time_notes} to correspond to the time mesh of the numerical solver, but rather consider the possibility of larger time steps which in many cases are advantageous for learning. We do however require that the nodes are equally spaced so $t_n=nT/N$. As a general rule, the number of time nodes $N$ is chosen based on the time resolution of the observed data and knowledge of the speed with which the physical system evolves. Let 
\begin{equation}
\u_n \in \R^m, \quad \z_n \in \R^s, \quad \text{and} \quad \w_n \in \R^q
\end{equation}
be the discretized state and parameters at time $t_n$. Here, $m$ should be thought of as high-dimensional since it corresponds to the nodes in the spatial mesh discretizing the state. The parameter dimensions $s$ and $q$ may be small or large depending upon the chosen parametrizations. We also note that $z$ and $w$ may be constant in time, but we have indexed them by time for generality. 

\section{Flow map learning} \label{sec:flow_map_learning}

We seek to build a surrogate for the flow map, defined as the function
\begin{eqnarray*}
\F:\R^m \times \R^s \times \R^q \to \R^m
\end{eqnarray*}
which evolves the state from time $t_n$ to $t_{n+1}$, i.e. 
\begin{equation}\label{eq:flow_map}
\u_{n+1} = \F(\u_n,\z_n,\w_n).
\end{equation}
This is in contrast to approaches that seek to learn a representation of the dynamical system, i.e., the state's time derivative. Rather, the flow map in \eqref{eq:flow_map} corresponds to the integral of the state's time derivative over an interval of length $\Delta t$. We specify a model for $\F$ in the form of a DNN and seek to train the parametrization given an ensemble of training and validation source-state pairs obtained from the computational model \eqref{eqn:dyn_pde}. To reduce the size of the DNN and facilitate training, we introduce dimension reduction described in Section \ref{sec:pca}. We then specify the architecture and training algorithm in Section \ref{sec:architecture_and_training}. 

\subsection{Ensemble of Simulation Data for Training and Validation}\label{sec:ensemble}
Assume that $M$ samples of $\mathbf{z}$ and $\mathbf{w}$ are given, which may either be sampled from a probability distribution or chosen based on domain knowledge. Denote them as
\begin{eqnarray*}
\z_n^i \in \R^s, \qquad \w_n^i \in \R^q, \qquad i=1,2,\dots,M,  \qquad n=0,1,\dots,N.
\end{eqnarray*}
Integrating the model~\eqref{eqn:dyn_pde} with these parameters yields a set of $M$ time series
\begin{eqnarray*}
\left\{ \u_n^{i} \right\}_{n=0}^N \subset \R^m, \quad i=1,2,\dots,M.
\end{eqnarray*}
Given that the generalization error of learned models is highly dependent on the training data, a foundational component of our framework is judicious sampling over a plausible range of parameters $(\z_n^i, \w_n^i)$.
We discuss the sampling of parameter space for generating training and validation data for our exemplar atmospheric dispersion problem in Section \ref{sec:flow_map_training}.

\subsection{Spatial Dimension Reduction of State and Parameters}\label{sec:pca}
The high dimensionality of $\u_n^{i} \in \R^m$ poses significant challenges to learning $\F$, as the model specification of $\F$ may result in an enormous number of parameters. As alluded to in Section \ref{sec:intro}, this is an inherent problem to operator learning in general, and most methods address it by reducing the dimensions of the input and output and by imposing structure in the operator model. Authors such as \citet{patel2021physics} and \citet{li2020neural} utilize Fourier bases to do so. We follow the ideas of of Operator Inference~\citep{opinf_peherstorfer_2016,opinf_mcquarrie_2021}, to project the data onto a linear solution manifold and learn the reduced dynamics in the lower-dimensional linear subspace. This form of linear dimension reduction is known as principal component analysis (PCA), proper orthogonal decomposition, or empirical orthogonal functions, depending on the area of literature~\citep{brunton2022data}, and was utilized by \citet{bhattacharya2020model} and \citet{hesthaven_2018} for operator learning. Prior to the dimension reduction, we center the data by subtracting the mean 
\begin{align*}
\mathbf{\mu}_{\u} = \frac{1}{N M} \sum\limits_{i=1}^{M} \sum\limits_{n=0}^N \u_n^{i}
\end{align*}
from each data vector~\citep{shlens2014tutorial}. For simplicity, we will abuse notation by using $\u_n^{i}$ to denote the data $\u_n^{i}-\mathbf{\mu}_{\u}$ after centering. All subsequent computations will be performed in the centered data space.

To determine a lower-dimensional linear subspace, the state's time snapshots are stacked together into a matrix
\begin{align*}
\mathbf{Y} = \left(
\begin{array}{cccccccccccc}
\u_0^{1} & \u_1^{1} & \cdots & \u_N^{1} & \u_0^{2} & \cdots & \u_N^{2} & \cdots & \u_0^{M} & \cdots & \u_N^{M}
\end{array}
\right) 
\in \mathbb{R}^{m \times (N+1)M}
\end{align*}
and its singular value decomposition $\mathbf{Y} =\U \mathbf{\Sigma} \mathbf{V}^\top$ is computed. Given an error threshold, we determine a rank $r \ll m$ for truncation and take the leading $r$ left singular vectors to define the linear subspace. Letting $\U_r \in \R^{m \times r}$ denote the leading $r$ columns of $\U$, left multiplication by $\U_r^\top \in \R^{r \times m}$ is referred to as PCA projection, while left multiplication by $\U_r$ is referred to as PCA reconstruction. We then seek to learn the \emph{reduced flow map}
\begin{align}\label{eq:reduced_flow_map_mapping}
\F_{r}:  \, \R^r \times \R^s \times \R^q &\to \R^r 
\end{align}
defined by
\begin{equation}\label{eq:reduced_flow_map_equation}
\F_r(\cc_n,\z_n,\w_n) = \U_r^\top \F(\U_r \cc_n,\z_n,\w_n),
\end{equation}
where 
\begin{equation}\label{eq:pca_proj}
\cc_n = \U_r^\top \u_n \in \R^r
\end{equation}
are the state variable coordinates in the reduced space defined by the columns of $\U_r$. 
Hereafter, we refer to $\cc_n$ as \emph{reduced state coordinates} for brevity. 

The idea of our approach is that if the state trajectory $\{\u_n\}_{n=0}^N$ is approximately contained in the range of $\U_r$ then the following diagram approximately commutes:
\[
\begin{tikzcd}[column sep = huge]
{\u_n} \arrow[r, "{\mathcal{F}(\cdot, \z_n, \w_n)}"] \arrow[d, "{\text{PCA projection} \, = \, \U_r^{\top}}"']
&
{\u_{n+1}}
\\
{\cc^n} \arrow[r, "{\mathcal{F}_{r}(\cdot, \z_n, \w_n)}"']
&
{\cc_{n+1}}
\arrow[u, "{\text{PCA reconstruction} \, = \, \U_r}"']
\end{tikzcd}
\]
However, the PCA projection \eqref{eq:pca_proj} is lossy in the sense that $\u_n \ne \U_r \cc_n = \U_r \U_r^T \u_n$, in general. The down-right-up path through the above diagram may be written as
\begin{align*}
\U_r \F_r(\U_r^\top \u_n,\z_n,\w_n) = \U_r \U_r^\top \F(\U_r \U_r^\top \u_n,\z_n,\w_n),
\end{align*}
which by \eqref{eq:flow_map}, approximates $\u_{n+1}$ up to the PCA truncation error resulting from the projector $\U_r \U_r^\top$. 

A guiding principle is to take $r$ large enough to ensure
\begin{equation}
\u_n^{i} \approx \U_r \U_r^T \u_n^{i}
\end{equation} 
for all of the training data $\{\u_n^i\}_{n=0}^N$, $i=1,2,\dots,M$. Taking a large $r$ results in $\mathcal{F}_r$ being a higher-dimensional operator and hence there is a tradeoff between the PCA truncation error $\u_n - \U_r \U_r^T \u_n$ and the error associated with approximating $\mathcal{F}_r$, discussed below in Subsection~\ref{sec:architecture_and_training}. Furthermore, the operator approximation will only be valid for trajectories which are approximately contained in the range of $\U_r$.

The input to the flow map $\mathcal{F}$ includes the parameters $\z_n$ and $\w_n$. If these are also high-dimensional, they present the same issue for learning the flow map as would the state variable. In the atmospheric dispersion example of Section \ref{sec:numerics}, $\w_n$ represents a wind field which has the same high dimension as the state $\u_n$. In such cases, dimension reduction of the parameters is required; we will apply the same PCA dimension reduction discussed above to $\w_n$, with a different rank than for the state $\u_n$. Once the parameters $\w_n$ and/or $\z_n$ in the training data have been reduced, the flow map $\mathcal{F}$ in the above framework may be considered a function of their reduced coordinates. Thus, while reduced coordinates may be used in place of the $\z_n$ and/or $\w_n$ data, for simplicity we do not introduce distinct notation for their reduced coordinates. 

Finally, we note that nonlinear dimension reduction approach have been proposed by, e.g., \citet{lee2020model} and \citet{barnett2022quadratic}. We focus on linear approaches as nonlinear methods may require larger training datasets that are more difficult to afford in many large-scale scientific applications. 

\subsection{Flow map approximation architecture and training}\label{sec:architecture_and_training}
We seek to learn a DNN approximation of $\F_r$ using the $MN$ training data pairs in reduced space
\begin{eqnarray*}
\left\{ (\cc_n^{i}, \z_n^{i},\w_n^{i}),\cc_{n+1}^{i} \right\} \qquad n=0,1,\dots,N-1, \qquad i=1,2,\dots,M,
\end{eqnarray*}
where $\cc_n^{i} = \U_r^\top \u_n^{i} \in \R^r$ are the reduced state coordinates. The previous subsection described how spatial structure of the state is embedded in the basis $\U_r$. Since the flow map is a time evolution operator, incorporating the temporal structure during training is critical to ensure stability of long-time, multi-step predictions. 
Consider DNN approximations $\N \approx \F_r$ of the form
\begin{align*}
 \N: \, & \R^r \times \R^s \times \R^q \times \R^\ell  \mapsto \R^r \\
 & (\cc,\z,\w;\bfxi)  \mapsto \cc + \Delta t \E(\cc,\z,\w;\bfxi)
\end{align*}
where $\E(\cc,\z,\w;\bfxi)$ is a dense feedforward DNN with weights and biases $\bfxi \in \R^\ell$ and $\Delta t=T/N$ is the time step size in the training data. This may interpreted as a forward Euler discretization of the governing PDE where $\E$ approximates the integral of the dynamics over a time window of length $\Delta t$ \citep{raissi2019physics, patel2022error}. 

\begin{figure}[htpb]
    \centering
    \includegraphics[width=0.7\textwidth]{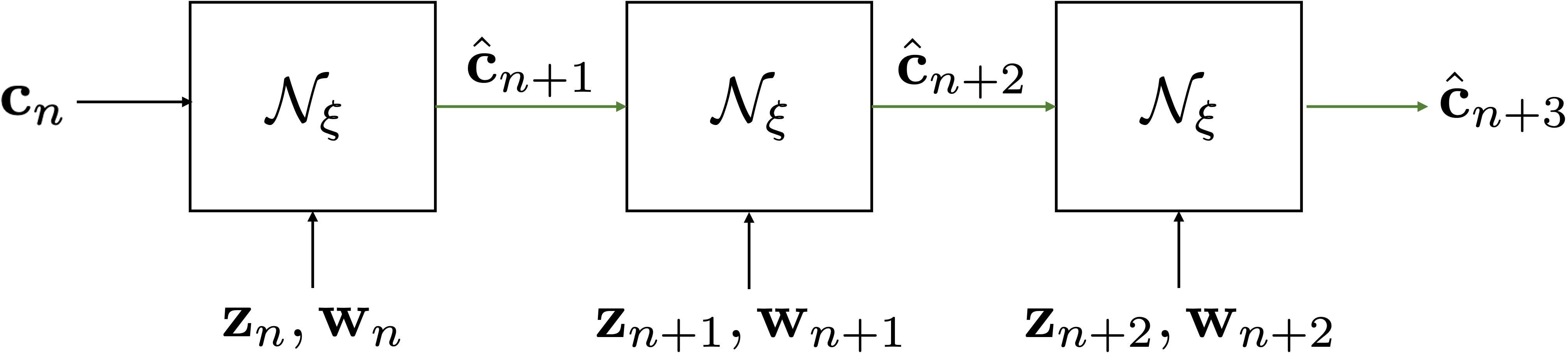}
    \caption{Example DNN composition for $p=3$, illustrating repeated composition through three time steps with relevant input and output labeled. The variable $\hat{\cc}_{n+3}$ represents the composed DNN estimate at time step $n+3$.} 
    \label{fig:example_P}
\end{figure}

The simplest $\ell_2$ loss function for training would involve summing over the squared error in the predictions for each data pair $\left\{ (\cc_n^{i}, \z_n^{i},\w_n^{i}),\cc_{n+1}^{i} \right\}$. Following \citet{patel2022error}, we include an additional sum over $P$ future time steps where we compose $\N$ with itself $P$ times. Specifically, we utilize the loss function
\begin{eqnarray}
\label{eqn:loss}
\L(\bfxi) = \sum\limits_{i=1}^{M} \sum\limits_{n=0}^{N-1} \sum\limits_{p=1}^{\overline{P}(n)} \| \cc_{n+p}^{i} - \N^{[p]}(\cc_n^{i},\{\z_j^i, \w_j^i\}_{j=n}^{n+p-1},\bfxi) \|_{\ell^2}^2
\end{eqnarray}
where $\N^{[p]}(\cc_n^{i},\{\z_j^i, \w_j^i\}_{j=n}^{n+p-1},\bfxi)$ denotes the composition of $\N$ with itself $p$ times, specifically
\begin{eqnarray}
\label{eqn:NNp}
\N^{[p]}(\cc_n,\{\z_j, \w_j\}_{j=n}^{n+p-1},\bfxi)
=
\N(\cdot, \z_{n+p-1}; \w_{n+p-1}; \bfxi) \circ \hdots 
\circ 
\N(\cdot, \z_{n};     \w_{n};     \bfxi) (\cc_n) 
\end{eqnarray}
where the $i$ superscript is omitted for readability. The interior sum from $p=1,\dots,\overline{P}(n)$ is truncated at $\overline{P}(n)=\min\{P,N-n\}$ to avoid predicting time steps outside the training data. The network composition with $p=3$ is illustrated in Figure \ref{fig:example_P}.
The hyperparameter $P$ dictates stability properties of the long term time integration using the composed flow map. In principle, the strongest enforcement of stability would require us to set $P=N$, thus incorporating all time step information into the training. However, increasing $P$ results in longer training times as it requires more network evaluation (which can only be partially batched) per loss function evaluation. Our numerical studies demonstrated a diminishing return on increasing $P$, but the necessity to take a sufficiently large $P$ to ensure stability and improve robustness in network training.

\section{Bayesian inversion using the learned flow map}\label{sec:bayes_inv}
Our ultimate goal is solving an inverse problem to estimate $\z$ given sparse and noisy measurements of $\u$. 
Let $\d_n \in \R^L$, $n=1,2,\dots,N$, denote observations of the state at $L$ spatial locations for each time step and $\mathcal O:\R^m \to \R^L$ denote the observation operator where $\d_n= \mathcal O(\u_n) + \mathbf{\epsilon}_n$, for noise vectors $\mathbf{\epsilon}_n$, $n=1,2,\dots,N$. We model $\{\mathbf{\epsilon}_n\}_{n=1}^N$ as independent identically distributed random vectors which follow a mean zero Gaussian distribution. Observation data may have more complex noise structure, but this is a common assumption when knowledge of the data noise is limited. 

In practice, the number of spatial observations $L$ is small relative to the full state dimension $m$. This causes the inverse problem to be ill-posed in the sense that there are many different values of $\z$ which are consistent with the observed data $\{\d_n\}_{n=1}^N$. To address this challenge, we consider a Bayesian formulation of the inverse problem which enables a systematic integration of prior knowledge, model predictions, and observed data to estimate $\z$. Furthermore, the Bayesian posterior distribution enables rigorous uncertainty quantification. In this section, we outline our Bayesian formulation of the inverse problem.

\subsection{Bayesian posterior distribution}
In this preliminary formulation of the Bayesian inverse problem we assume that some prior knowledge is available to constrain the range of possible $\z$'s, such as a range of possible magnitudes or knowledge of length scales which characterize its smoothness. This information is embedded in a prior distribution. As is common~\citep{smith2013uncertainty}, we assume that this prior is Gaussian with mean $\overline{\z}$ and covariance $\mathbf{\Gamma}_{\text{prior}}$, and let $\pi_{\text{prior}}(\z)$ denote its probability density function (PDF). Recalling our assumption of additive Gaussian noise $\mathbf{\epsilon}_n$ contaminating the data, we have a Gaussian likelihood $\pi_{\text{like}}(\z \vert \d)$ which results from evaluating the noise model PDF at the observed data 
\begin{equation}
\d=(\d_1,\d_2,\dots,\d_N) \in \R^{LN}.
\end{equation}
Bayes' Theorem gives the posterior distribution of $\z$ as
\begin{eqnarray}
\label{eqn:post_pdf}
\pi_{\text{post}}(\z) \propto \pi_{\text{prior}}(\z) \pi_{\text{like}}(\z \vert \d)
\end{eqnarray}
where we have omitted the normalizing constant which is not needed in our analysis. The point of greatest posterior probability is called the maximum a posteriori probably (MAP) point and gives a best estimate of $\z$. Drawing samples from the posterior distribution quantifies uncertainty about the MAP due to limitations on data availability and quality. 

\subsection{Bayesian approximation error}
The Bayesian posterior implicitly depends on $\mathbf{w}$ which is uncertain. We assume that a probabilistic model (for which samples may be computed) for $\w$ is given. Let $\overline{\w}$ denote the best point estimate of $\w$, for instance, the mean of the probabilistic model. We use the Bayesian approximation error (BAE) approach to embed uncertainty from $\w$ into the inverse problem estimating $\z$ ~\citep{nicholson_2018,Arridge_2006,Kaipio_2013}.

Let 
\begin{equation}\label{eq:params_to_obs}
\mathbf{F}_n(\z,\w)=\mathcal O \left( \U_r \N^{[n]}(\U_r^\top \u_0,\{\z_j,\w_j\}_{j=0}^n,\bfxi) \right)
\end{equation}
denote the mapping from $(\z,\w)$ to the observations of the state at time step $t_n$, defined by composing the flow map approximation with itself $n$ times and starting from the initial condition $\u_0$. The parameter-to-observable map is defined by concatenating $\mathbf{F}_n(\z,\w)$ for all time steps,
\begin{align*}
\mathbf{F}(\z,\w) = 
\left(
\begin{array}{c}
\mathbf{F}_1(\z,\w) \\
\mathbf{F}_2(\z,\w) \\
\vdots \\
\mathbf{F}_N(\z,\w)
\end{array}
\right) \in \R^{LN} .
\end{align*}
A common practice merely fixes $\w=\overline{\w}$ to its best estimate and solves the inverse problem for $\z$, i.e. it uses the parameter-to-observable map $\mathbf{F}(\z,\overline{\w})$. This is necessary in many applications where the data is not sufficiently informative to simultaneously estimate both $\z$ and $\w$, but it fails to incorporate uncertainty in $\w$ and may provide poor uncertainty quantification.  

Our previous assumption of Gaussian noise implies that $\d = \mathbf{F}(\z,\w^\star) + \mathbf{\epsilon}$
where $\w^\star$ is the true (but unknown) value of $\w$ and $\mathbf{\epsilon}$ is a $NL$-dimensional noise vector (the concatenation of $\{ \mathbf{\epsilon}_n\}_{n=1}^N$) which we assumed to be Gaussian with mean zero and covariance $\mathbf{\Gamma}_{\text{noise}}$. We assume the noise covariance has been specified to reflect knowledge of the data and/or model fidelity. Since $\w^\star \ne \overline{\w}$, we incorporate $\w$'s uncertainty by considering
\begin{align*}
\d = \mathbf{F}(\z,\w^\star) + \mathbf{\epsilon} = \mathbf{F}(\z,\overline{\w}) + \mathbf{\epsilon} + (\mathbf{F}(\z,\w^\star)-\mathbf{F}(\z,\overline{\w}))  = \mathbf{F}(\z,\overline{\w}) + \mathbf{\nu}.
\end{align*}
Here, the random vector $\mathbf{\nu}=\mathbf{\epsilon} + (\mathbf{F}(\z,\w^\star)-\mathbf{F}(\z,\overline{\w}))$ models the combination of data noise and the model error arising from misspecification of $\w$. Since $\w^\star$ is unknown, the distribution of $\mathbf{\nu}$ must be approximated. We consider the random vector $\mathbf{e}=\mathbf{F}(\z,\w)-\mathbf{F}(\z,\overline{\w})$ corresponding to the push forward of the prior distribution for $\z$ and the probability distribution modeling the uncertainty in $\w$, which may be thought of as a prior. Due to potential nonlinearities, the distribution of $\mathbf{e}$ may not be Gaussian and in general is difficult to characterize. However, it is computationally advantageous and common in practice to adopt a Gaussian approximation of it. To this end, we generate samples $\z_1,\z_2,\dots,\z_b$ and $\w_1,\w_2,\dots,\w_b$ of $\z$ and $\w$, respectively, and evaluate the parameter-to-observable map for the $b$ samples. The model error statistics are approximated as a Gaussian with an empirical mean and covariance  
\begin{align*}
\overline{\mathbf{e}} = \frac{1}{b} \sum\limits_{\ell=1}^b \mathbf{e}_\ell \qquad \text{and} \qquad \mathbf{\Gamma}_{\mathbf{e}} = \frac{1}{b-1}\sum\limits_{\ell=1}^b ( \mathbf{e}_\ell - \overline{\mathbf{e}}) ( \mathbf{e}_\ell - \overline{\mathbf{e}})^\top
\end{align*}
where $\mathbf{e}_\ell = \mathbf{F}(\z_\ell,\w_\ell)-\mathbf{F}(\z_\ell,\overline{\w})$. Then, rather than the noise model $\mathbf{\epsilon}$ with mean zero and covariance $\mathbf{\Gamma}_{\text{noise}}$, we use the noise model $\mathbf{\nu}$ with mean $\overline{\mathbf{e}}$ and covariance 
\begin{equation}
\mathbf{\Gamma}_{\text{BAE}} = \mathbf{\Gamma}_{\text{noise}}+\mathbf{\Gamma}_{\mathbf{e}}.
\end{equation}

This results in a likelihood which includes a data bias corresponding to the model error mean $\overline{\mathbf{e}}$ and a larger noise covariance which is the sum of the data noise and model error covariances. By incorporating explicit knowledge of how uncertainty in $\w$ propagates through the model, we inform the inverse problem of how closely the model should seek to match the observational data. This aids to reduce overfitting and encourages better uncertainty quantification. 

\subsection{MAP point estimation}
To determine the MAP point, we may formulate an optimization problem to maximize the posterior PDF. Rather than solving this problem directly, it is numerically advantageous to minimize the negative log of the posterior PDF, the minimizer of which coincides with the MAP point since the logarithm is a monotonic function. Specially, we solve the optimization problem
\begin{align}
\label{eqn:map_point_opt}
\mn_{\z} 
\left\{
J(\z) \coloneqq \frac{1}{2} \sum\limits_{n=1}^N \| \mathbf{F}_n(\z) + \overline{\mathbf{e}}  - \d_n \|_{\mathbf{\Gamma}_{\text{BAE}}^{-1}}^2 + \frac{1}{2} \| \z - \overline{\z} \|_{\mathbf{\Gamma}_{\text{prior}}^{-1}}^2
\right\}.
\end{align}
The subscripts $\mathbf{\Gamma}_{\text{BAE}}^{-1}$ and $\mathbf{\Gamma}_{\text{prior}}^{-1}$ denote norms weighted by the inverse of the covariance matrices, also known as the Mahalanobis norm \citep{maupin2018validation}.

To solve~\eqref{eqn:map_point_opt} efficiently we use derivative-based optimization algorithms which use the gradient of $J(\z)$ to determine the optimization steps. Further, our posterior sampling will utilize the Gauss-Newton Hessian approximation. In what follows, we describe how the gradient and Gauss-Newton Hessian may be efficiently computed using our flow map approximation. This simplifies the inverse problem computation by leveraging derivative information to traverse the space of possible $\z$'s without requiring extensive tuning of algorithmic parameters.

First, we observe that the objective function \eqref{eqn:map_point_opt} can be decomposed as 
\begin{equation}
J(\z)=M(\z)+R(\z),
\end{equation}
where $M(\z)$ is the misfit term involving the model prediction and data, and $R(\z)$ is the quadratic regularization term corresponding to the negative log of the prior PDF. The gradient and Hessian of $R(\z)$ are easily computed so we focus on the misfit $M(\z)$, which requires differentiating through the flow map approximation. 

To evaluate $M(\z)$, its gradient $\nabla M(\z)$, and its Gauss-Newton Hessian approximation $\nabla_{\text{GN}}^2 M(\z)$, we assume that $\w$, $\u_0$, and $\bfxi$ are given and fixed. The chain rule and product rule give
\begin{align}
\label{eqn:misfit_grad}
\nabla M(\z) = \sum\limits_{n=1}^N \frac{\partial \mathbf{F}_n}{\partial \z}(\z) \mathbf{\Gamma}_{\text{BAE}}^{-1} \left( \mathbf{F}_n(\z) + \overline{\mathbf{e}}   - \d_n \right).
\end{align}
Differentiating the misfit gradient~\eqref{eqn:misfit_grad} with respect to $\z$ yields
\begin{align}
\nabla^2 M(\z) = \sum\limits_{n=1}^N \left( \frac{\partial \mathbf{F}_n}{\partial \z}(\z) \right)^\top \mathbf{\Gamma}_{\text{BAE}}^{-1} \frac{\partial \mathbf{F}_n}{\partial \z}(\z)  +  \frac{\partial^2 \mathbf{F}_n}{\partial \z^2}(\z) \mathbf{\Gamma}_{\text{BAE}}^{-1} \left(  \mathbf{F}_n(\z) +\overline{\mathbf{e}}  - \d_n  \right)
\end{align}
where
$ {\partial^2 \mathbf{F}_n} / {\partial \z^2}$
denotes the three-dimensional tensor corresponding to the second derivative of the vector-valued function $\mathbf{F}_n$. This second derivative is difficult to compute, so the Gauss Newton Hessian approximation omits it:
\begin{align}
\nabla_{\text{GN}} ^2 M(\z) = \sum\limits_{n=1}^N \left( \frac{\partial \mathbf{F}_n}{\partial \z}(\z) \right)^\top \mathbf{\Gamma}_{\text{BAE}}^{-1} \frac{\partial \mathbf{F}_n}{\partial \z}(\z)  .
\end{align}
Such an omission is not justified in general; however, since $\mathbf{F}_n(\z) + \overline{\mathbf{e}}   - \d_n $ will be small for $\z$'s near the minimizer, i.e., where the model fits the data, we have $\nabla_{\text{GN}} ^2 M(\z) \approx \nabla^2 M(\z)$ around the MAP point. 

To efficiently compute $M(\z)$, $\nabla M(\z)$, and $\nabla_{\text{GN}}^2 M(\z)$, we may use algorithmic, or automatic, differentiation at each time step when evaluating the flow map approximation. Specifically, for a given $\z$, we loop over $n=1,2,\dots,N$, to evaluate $\N$, 
${\partial \N}/{\partial \cc_n}$, 
and
${\partial \N}/{\partial \z_n}$
at each time step. This can be done using automatic differentiation algorithms in various deep learning platforms. We have utilized TensorFlow 2 for the results in this article, and made use of the \texttt{tf.GradientTape} feature. By \eqref{eq:params_to_obs}, the Jacobian of the prediction at time step $k$, 
${\partial \mathbf{F}_k}/{\partial \z}$,
depends on the Jacobians of $\N$ at the current time step and at each previous time step $t_n$, $n=1,2,\dots,k$. Hence, we compute and store the evaluations of the Jacobians of $\N$ for each time step. Then, we recursively access them to calculate matrix-vector products with ${\partial \mathbf{F}_n}/{\partial \z}$ needed in the gradient
and in the Gauss-Newton Hessian vector product $\nabla_{\text{GN}} ^2 M(\z) \mathbf{v}$. 
Notably, we never form large matrices, but rather form and store matrices for each time step which are accessed recursively to compute the necessary global derivative information. By providing accurate derivative information, we are able to solve~\eqref{eqn:map_point_opt} with state-of-the-art derivative-based optimization algorithms which exhibit accelerated convergence. We utilized the Rapid Optimization Library \citep{ROL-website}, part of the Trilinos package \citep{trilinos-website} provided by Sandia National Laboratories. 

\subsection{Posterior sampling}
Computing the MAP point, which we denote as $\z_{\text{MAP}}$, provides a best estimate of $\z$. However, it does not characterize uncertainty in the estimate. To quantify uncertainty, we seek to compute samples from the posterior distribution. However, this may be challenging for high-dimensional problems. To facilitate rapid sampling, we employ the Laplace approximation of the posterior distribution. This assumes that $\mathbf{F}_n(\z)$ is linear in $\z$, which is valid in a neighborhood of the MAP point. Under such assumptions of linearity, and our assumptions of a Gaussian prior and noise model, the posterior distribution is Gaussian with its mean given by the MAP point and its covariance given by the inverse Hessian evaluated at the MAP point, $\nabla^2 J(\z_{\text{MAP}})^{-1}$. Though the Laplace approximation fails to capture complex multi-modal posterior distributions, it provides a pragmatic way to estimate uncertainty.

To generate approximate posterior samples using the Laplace approximation, we recall that 
\begin{equation}
\nabla_{\text{GN}} ^2 J(\z_{\text{MAP}}) \approx \nabla^2 J(\z_{\text{MAP}}),
\end{equation}
where $\nabla_{\text{GN}} ^2 J = \nabla_{\text{GN}} ^2 M + \nabla^2 R$. Hence we seek to sample from the Gaussian distribution with mean $\z_{\text{MAP}}$ and covariance $\nabla_{\text{GN}} ^2 J(\z_{\text{MAP}})^{-1}$. Such sampling can be achieved by factorizing $\nabla_{\text{GN}} ^2 J(\z_{\text{MAP}})^{-1} = \mathbf{L} \mathbf{L}^\top$, and computing vectors of the form $\z_{\text{MAP}} + \mathbf{L} \mathbf{\omega}$, where the entries of the vector $\mathbf{\omega}$ are independent samples from a standard normal distribution. 

Since we generally only have access to $\nabla_{\text{GN}} ^2 J(\z_{\text{MAP}})$ via matrix-vector products, we avoid forming $\nabla_{\text{GN}} ^2 J(\z_{\text{MAP}})^{-1}$ explicitly or factorizing it directly. Rather, observe that
\begin{align*}
\nabla_{\text{GN}} ^2 J(\z_{\text{MAP}})^{-1} =  \left( \nabla_{\text{GN}} ^2 M(\z_{\text{MAP}}) + \mathbf{\Gamma}_{\text{prior}}^{-1} \right)^{-1}.
\end{align*}
Frequently, the inverse of the prior covariance admits a factorization (due to being defined as the square of a differential operator) for the form $\mathbf{\Gamma}_{\text{prior}}^{-1}= \mathbf{E} \mathbf{E}^\top$. Then
\begin{align*}
\nabla_{\text{GN}} ^2 J(\z_{\text{MAP}})^{-1} &=  \left( \nabla_{\text{GN}} ^2 M(\z_{\text{MAP}}) + \mathbf{E} \mathbf{E}^\top \right)^{-1} \\
&= \mathbf{E}^{-\top} \left( \mathbf{E}^{-1} \nabla_{\text{GN}} ^2 M(\z_{\text{MAP}}) \mathbf{E}^{-\top} + \mathbf{I} \right)^{-1} \mathbf{E}^{-1},
\end{align*}
where $\mathbf{I}$ is the identity matrix and $\mathbf{E}^{-\top}$ denotes the inverse transpose of $\mathbf{E}$. Given access to matrix-vector products with $ \nabla_{\text{GN}} ^2 M(\z_{\text{MAP}}) $, we compute a truncated eigenvalue decomposition
\begin{equation}
\label{eqn:hess_eig_trunc}
\mathbf{E}^{-1} \nabla_{\text{GN}} ^2 M(\z_{\text{MAP}}) \mathbf{E}^{-\top} \approx \mathbf{V}_k \mathbf{\Lambda}_k \mathbf{V}_k^\top
\end{equation}
of rank $k$. Then applying the Sherman-Morrison-Woodbury formula to $\mathbf{V}_k \mathbf{\Lambda}_k \mathbf{V}_k^\top + \mathbf{I}$ gives
\begin{align*}
\nabla_{\text{GN}} ^2 J(\z_{\text{MAP}})^{-1} \approx \mathbf{E}^{-1} \left( \mathbf{V}_k \mathbf{\Lambda}_k \mathbf{V}_k^\top + \mathbf{I} \right)^{-1} \mathbf{E}^{-\top} = \mathbf{E}^{-1} (\mathbf{I} - \mathbf{V}_k \mathbf{D}_k \mathbf{V}_k^\top )\mathbf{E}^{-\top},
\end{align*}
where $\mathbf{D}_k$ is a $k \times k$ diagonal matrix whose entries are $\lambda_i/(1+\lambda_i)$, $\lambda_i$ being the $i^{\text{th}}$ eigenvalue. The rank $k$ truncation error in~\eqref{eqn:hess_eig_trunc} is $\mathcal O(\lambda_{k+1})$ and hence is small for many inverse problem where data sparsity causes the misfit Hessian to be low rank.

Then $\nabla_{\text{GN}} ^2 J(\z_{\text{MAP}})^{-1} \approx \mathbf{L} \mathbf{L}^\top,$ where 
\begin{align*}
\mathbf{L} = \mathbf{E}^{-1} (\mathbf{I} + \mathbf{V}_k \mathbf{P}_k \mathbf{V}_k^\top ),
\end{align*}
and $\mathbf{P}_k$ is a $k \times k$ diagonal matrix whose entries are $-1+1/\sqrt{1+\lambda_i}$. Approximate posterior samples may be efficiently computed by applying $\mathbf{L}$ to standard normal Gaussian random vectors. This approach is common in the large-scale Bayesian inverse problems literature thanks to its computational efficiency and algorithmic simplicity; for a general reference on the material presented in this section, see~\citet{isaac2015scalable}. Note that other posterior sampling algorithms may be employed to avoid assuming that the posterior is approximately Gaussian. They were not needed in the numerical results presented below; however, many sampling algorithms exist which leverage derivative information to efficiently explore complex high dimensional posterior distributions. 

\section{Numerical results} \label{sec:numerics}
We now demonstrate our proposed approach on an idealized atmospheric aerosol dispersion model, for which we seek to infer a time varying magnitude function in an aerosol source. 

\subsection{Atmospheric Aerosol Dispersion Model}
Data is generated from the advection-diffusion-reaction PDE
\begin{align}
\label{eqn:adv_diff_reaction_pde}
& \frac{\partial u}{\partial t} - \kappa \nabla^2 u + \mathbf{v}(w) \cdot \nabla u - S \mathbf{e}_y \cdot \nabla u = \mathcal R(u) + f(z)& \text{on } \Omega \times (0,\infty) \\
& \nabla u \cdot \mathbf{n} = 0 & \text{on } \partial \Omega \times (0,\infty) \nonumber \\
& u = 0 & \text{on } \Omega \times \{0\} \nonumber
\end{align}
where the spatial domain $\Omega = [0,200] \times [0,20]$ represents longitude $(km)$ and altitude $(km)$ for a plume of $\text{SO}_2$ aerosols being emitted from a source $f$ for a $T=60$ minute duration of time. The boundary of $\Omega$ is denoted as $\partial \Omega$ and the boundary's outward facing normal vector is denoted as $\mathbf{n}$. The unit vector in the vertical direction is $\mathbf{e}_y=(0,1)$.

The source is defined as
\begin{align*}
f(z) = z(t)F(x,y)
\end{align*}
where $z:[0,T] \to \R$ is the time varying source magnitude being inferred and 
\begin{align*}
F(x,y)= \exp{ \left( -100(x-5)^2 \right) }  \exp{ \left( -0.1(y-9)^2 \right)}
\end{align*}
 is the spatial profile of the source which is assumed to be stationary and known. Hence, $\z$ will be a vector whose length is the number of time steps $N$ in the discretized data.

The advection velocity $\mathbf{v}=(w,0)$ has a zero vertical component and an uncertain horizontal component $w:\Omega \times [0,T] \to \R$ corresponding to the spatially heterogenous and temporally varying winds. The diffusion coefficient $\kappa$, terminal falling speed $S$, and reaction function $R$ are assumed to be known. The complete details of the model description are given in Appendix A.

\subsection{Flow Map Training}\label{sec:flow_map_training}

\subsubsection*{Data Generation} 
The discretized data has $N=120$ time nodes and $m=101101$ spatial nodes. Hence, $\z$ is represented by a vector in $\R^{120}$ and $\w$ is represented by a matrix in $\R^{101101 \times 120}$, or equivalently a vector in $\R^{(101101)(120)}$, when expressing it as a vector is more convenient. The state solution $\u$ has the same dimension as the wind $\w$. To train a flow map approximation, we generate a dataset by solving the PDE~\eqref{eqn:adv_diff_reaction_pde} for $M=16$ combinations of source magnitudes and wind fields.  These $(\z,\w)$ inputs are determined by choosing $4$ source magnitudes, $4$ wind fields, and taking all possible combinations of them to generate $16$ $(\z,\w)$ pairs. This follows common conventions in earth system simulation which use ensembles of climate states (which are represented by wind fields in our simplified example) and replicate the ensemble for each forcing scenario. The source magnitudes\footnote{The source magnitudes are generated by parameterizing as $z(t) = 3 \times 10^3 \left( \eta_1 \exp\left( - 0.015 \eta_2 t \right) \right)$ and taking the $(\eta_1,\eta_2)$ from the set $\{(2,0.5),(2,2),(3.5,0.5),(3.5,2)\}$.} are shown in the left panel of Figure~\ref{fig:source_data} with the label `training data'. These profiles were selected to capture a range of decay rates and initial source magnitudes.  

Wind field samples are generated by parameterizing $\w$ using a vector $\mathbf{\theta} \in \R^9$ and randomly sampling the components of $\mathbf{\theta}$ from uniform distributions.  
The parameterization of $\w$ and ranges of these distributions are specified in Appendix A. 
A mean wind $\overline{\w}$ is generated by evaluating the parameterized wind field at the mean of $\mathbf{\theta}$. 
To ensure that our wind field training data are sufficiently distinct, we generate many wind field samples and choose $4$ of them which are the farthest (in the $\ell^2$ vector norm) from the mean wind field $\overline{\w}$. For these samples $\{\w^k\}_{k=1}^4$, their relative $\ell^2$ distance from the mean
\begin{align*}
\frac{\| \w^k - \overline{\w}\|_{\ell^2}^2 }{\| \overline{\w}\|_{\ell^2}^2 },
\end{align*}
varies from $3.3\%$ to $3.7\%$. Later in this section we will increase the wind field variability to demonstrate how it affects the flow map approximation and resulting inverse problem. The top row of Figure~\ref{fig:wind_data} displays three time snapshots of the mean wind field to illustrate its spatiotemporal characteristics. 

Denoting the source magnitude samples as $\z^j$, $j=1,\dots,4$, wind field samples as $\w^k$, $k=1,\dots,4$, and state solutions as $\u^{j,k}$, $j=1,\dots,4$, $k=1,\dots,4$, we divide our dataset into a training dataset consisting of data from 12 PDE solves
\begin{align*}
\{ \z^j,\w^k,\u^{j,k} \} \qquad j=1,\dots,4, \qquad  k=1,\dots,3,
\end{align*}
 and a validation dataset consisting of 4 PDE solves
 \begin{align*}
 \{ \z^j,\w^4,\u^{j,4} \}, \qquad j=1,\dots,4.
 \end{align*}
 The validation set is used to tune the flow map architecture, as discussed in the next section. The final flow map using the tuned set of hyperparameters is trained on all 16 forward solves.  
 
 Using the approach outlined in Section \ref{sec:pca}, we compress the state with a rank $r=70$ PCA representation, which was determined by considering the tradeoff between both the reconstruction error and the computational cost of training larger DNNs.  The maximum relative ${\ell}_2$ reconstruction error for the selected PCA basis on the validation set was 0.6 \%. We also generate a lower-dimensional basis for the wind fields using $r=10$ modes with 0.7\% reconstruction error.  In both cases the PCA basis was generated using the training dataset. Since the source magnitude is scalar, the reduced flow map $\F_r$ given by \eqref{eq:reduced_flow_map_equation} maps $\mathbb{R}^{81}$ into $\mathbb{R}^{70}$.

\subsubsection*{Hyperparameter Tuning}
\begin{figure}[htpb]
    \centering
    \includegraphics[width=0.5\textwidth]{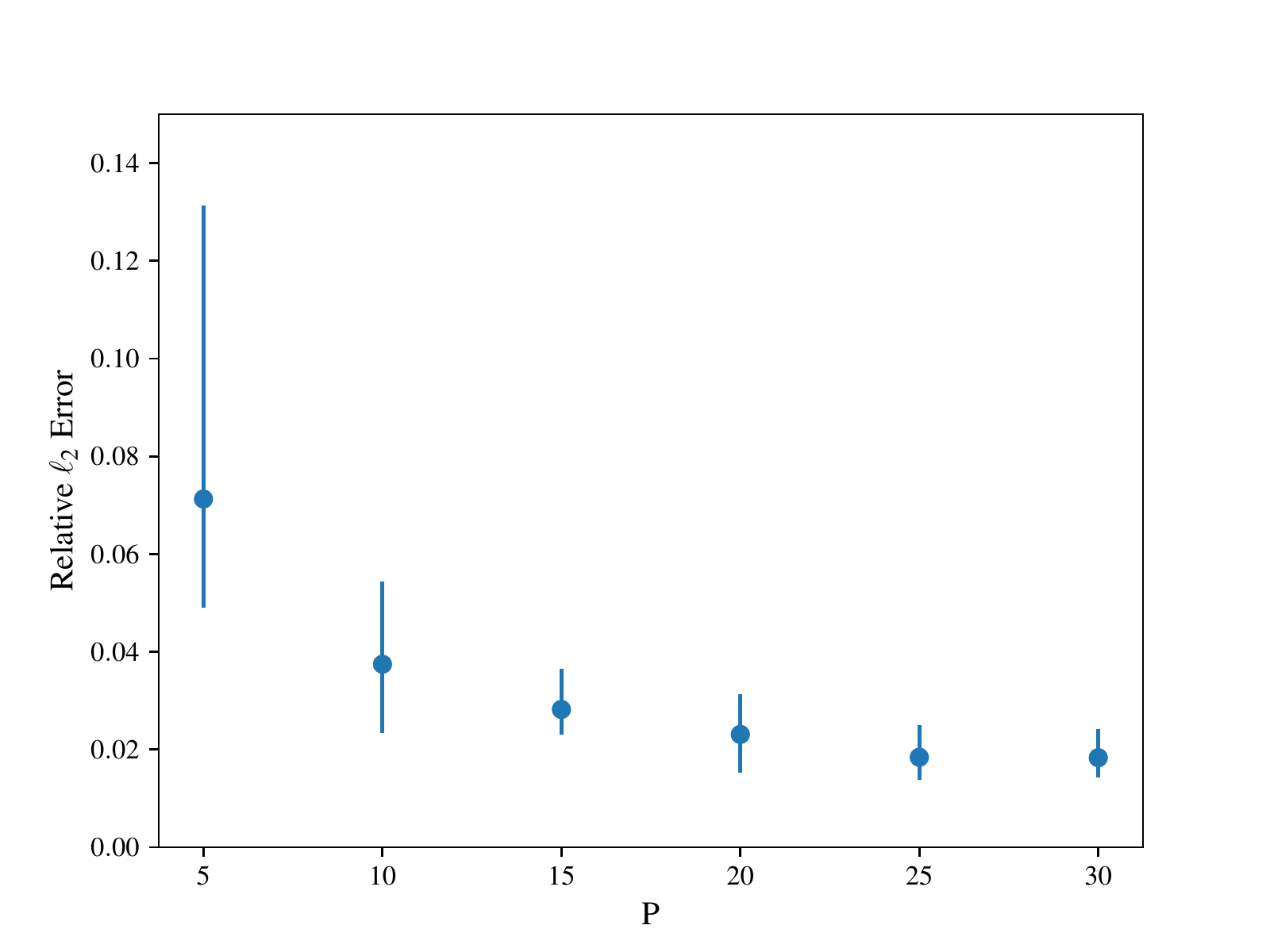}
    \caption{Impact of increasing $P$, the number of network compositions, on flow map performance.  Error bars indicate the minimum and maximum values across 6 training instances. } 
    \label{fig:tune_P}
\end{figure}

We tuned the DNN by comparing the relative $\ell_2$ error on the validation set following training for different DNN configurations. The relative $\ell_2$ error on the $M_\text{val}=4$ validation forward solves is computed as follows:

\begin{eqnarray}
\label{eqn:l2}
\frac{1}{M_\text{val} N} \sum\limits_{i=1}^{M_\text{val}} \sum\limits_{n=1}^{N} \frac{\| \u_n^i - \U_r \N^{[n]}(\cc_0^{i},\{\z_j^i, \w_j^i\}_{j=0}^{n-1},\bfxi) \|_{\ell^2}^2}{\| \u_n^i\|_{\ell^2}^2 },
\end{eqnarray}
where we emphasize again that error is measured by repeated composition of the flow map with itself rather than a time-step-wise error. For all experiments, we used the ADAM optimizer and ELU nonlinear activation layers.  We first examined the impact of increasing $P$, the number of DNN compositions used in the loss function~\eqref{eqn:loss}. For each value of $P$, we trained 6 flow maps, each using a random Glorot initialization \citep{glorot2010understanding}.  The relative $\ell_2$ scores on the validation set are shown in Figure \ref{fig:tune_P}, with error bars indicating the minimum and maximum $\ell_2$ values across training instances.  
We found that increasing $P$  increased the network accuracy as well as stability, with less variance across the 6 training instances for larger values of $P$.  Beyond 25 network compositions, there was limited benefit to further increasing $P$ in terms of accuracy, and larger values of $P$ were more computationally expensive. 
 
We also conducted a study to determine the optimizer learning rate, and the hidden layer width and depth. The selected final configuration uses width 200, depth 2, and learning rate 0.0008.  We detail the configurations used for all studies in Appendix B. 
For the final flow map which is used for the inversion study, we retrain the model on all 16 forward solves using $P=25$.

\subsection{Inversion}

\subsubsection*{Data generation}

\begin{figure}
    \centering
    \includegraphics[width=0.32\textwidth]{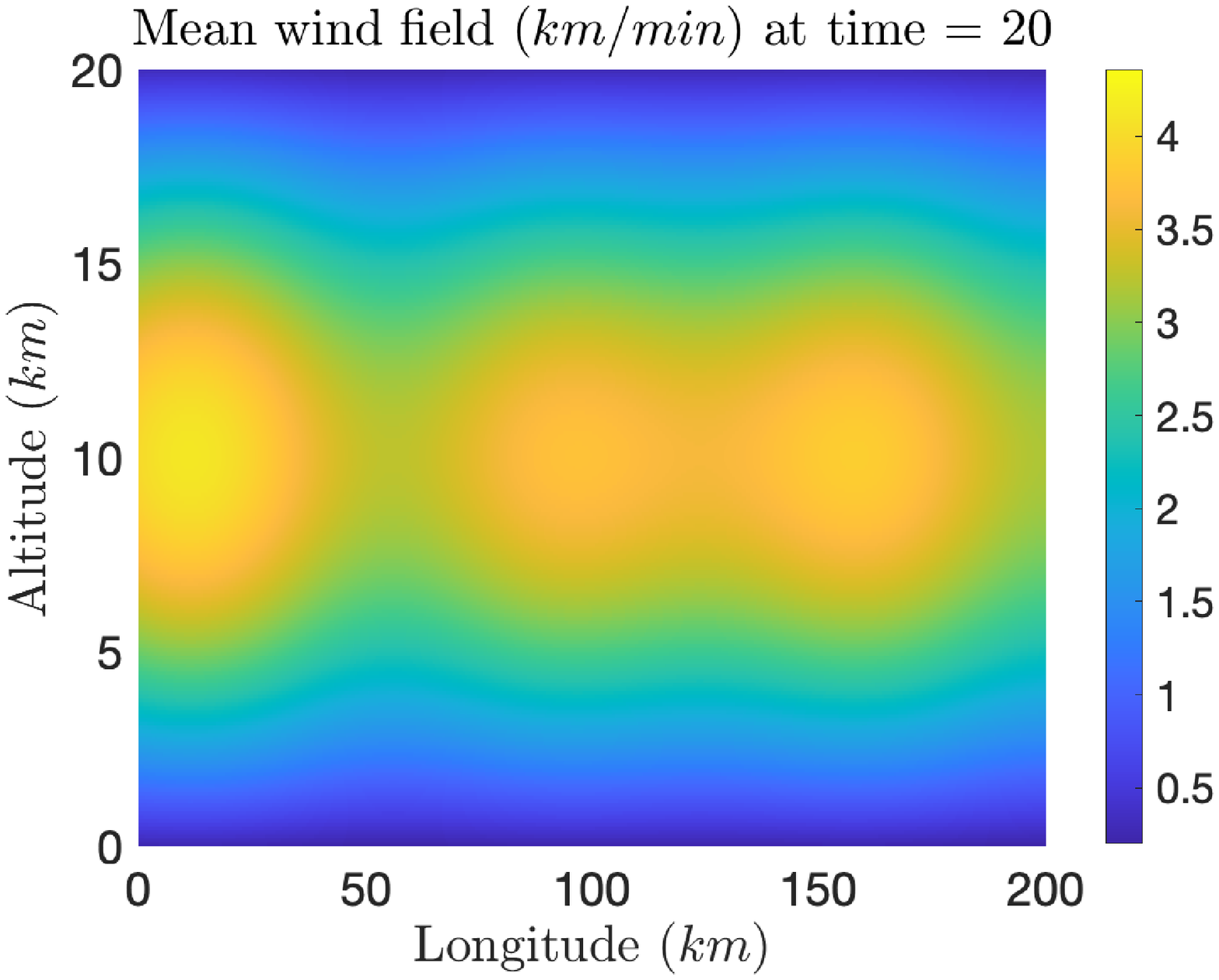}
    \includegraphics[width=0.32\textwidth]{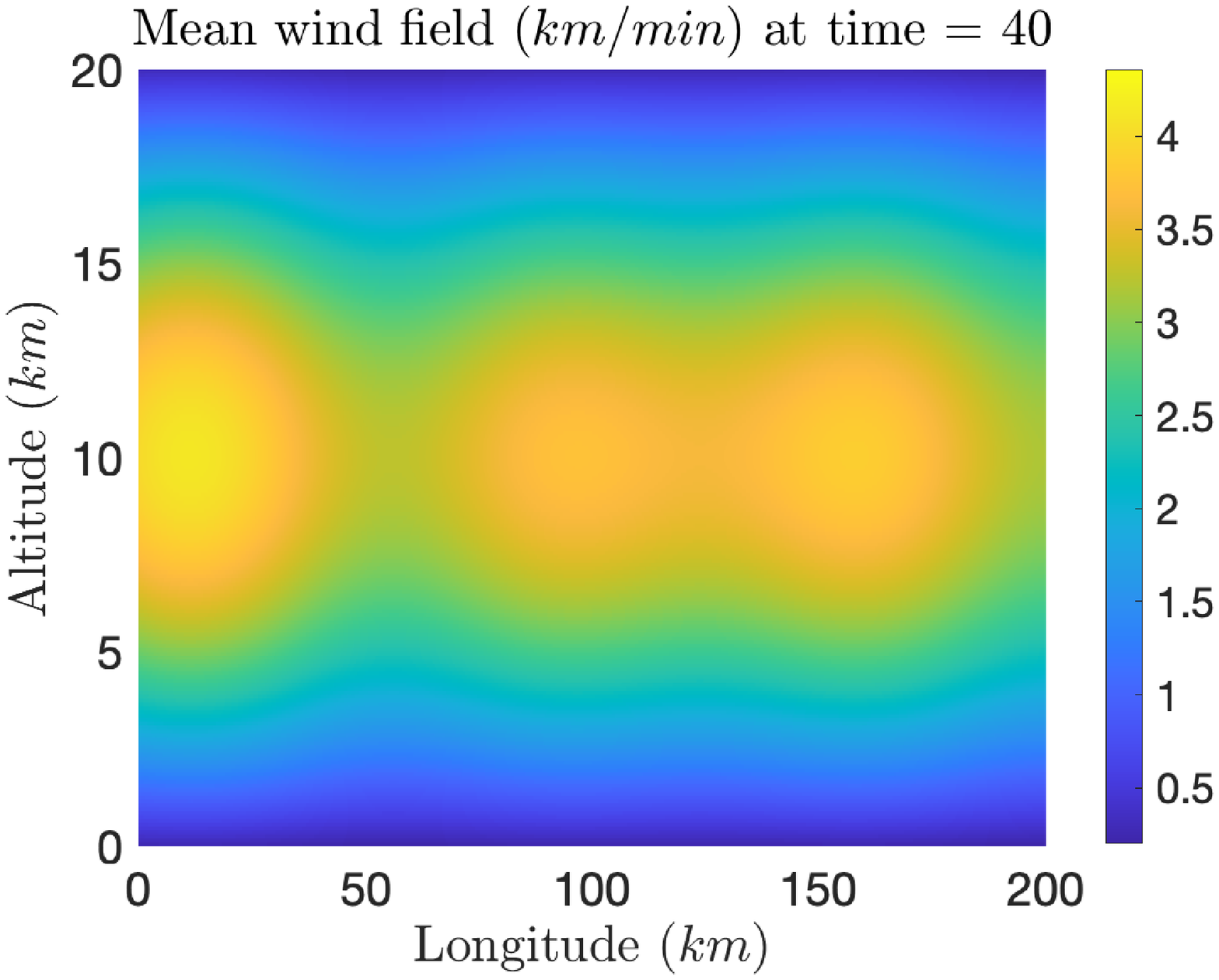}
    \includegraphics[width=0.32\textwidth]{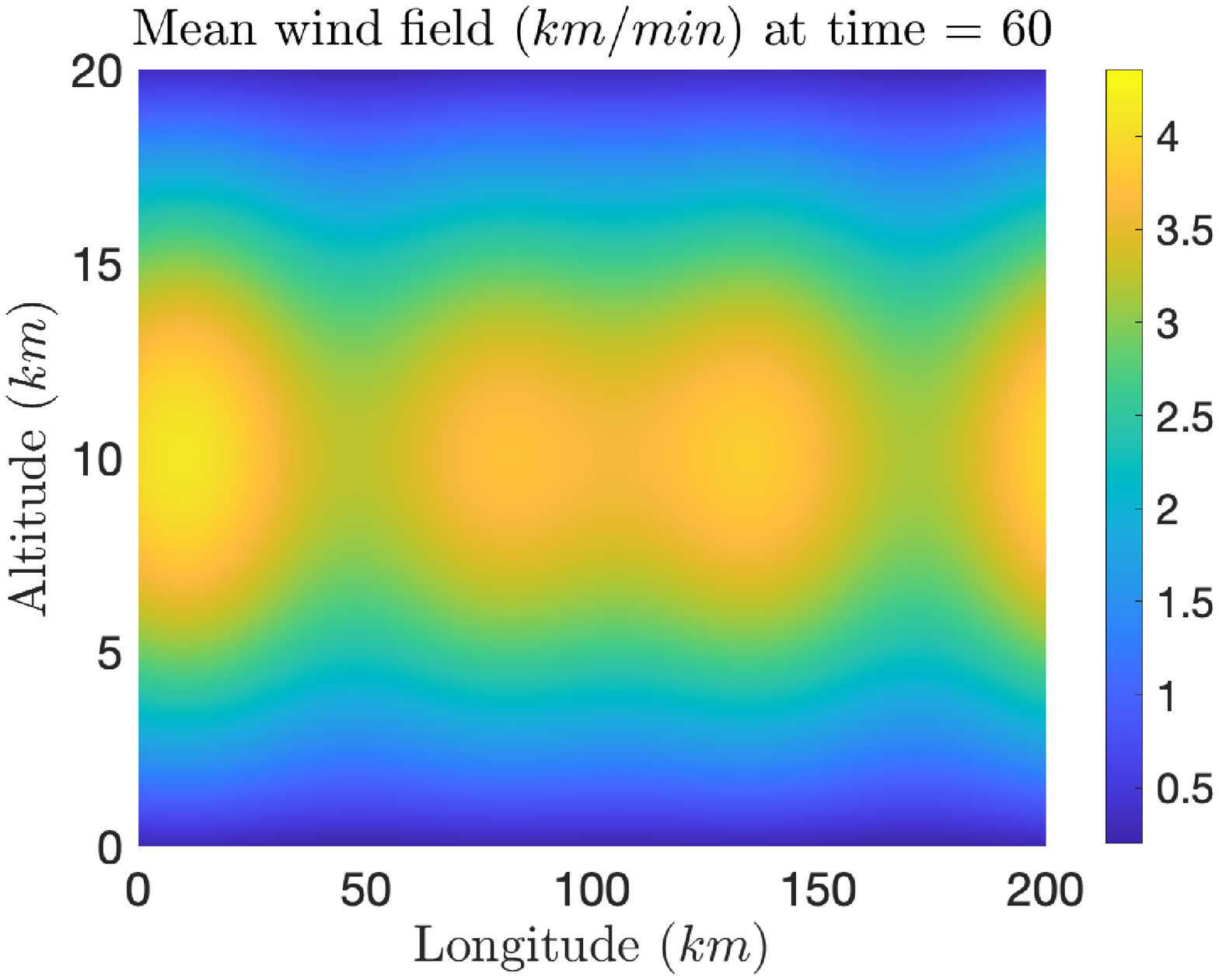} \\
        \includegraphics[width=0.32\textwidth]{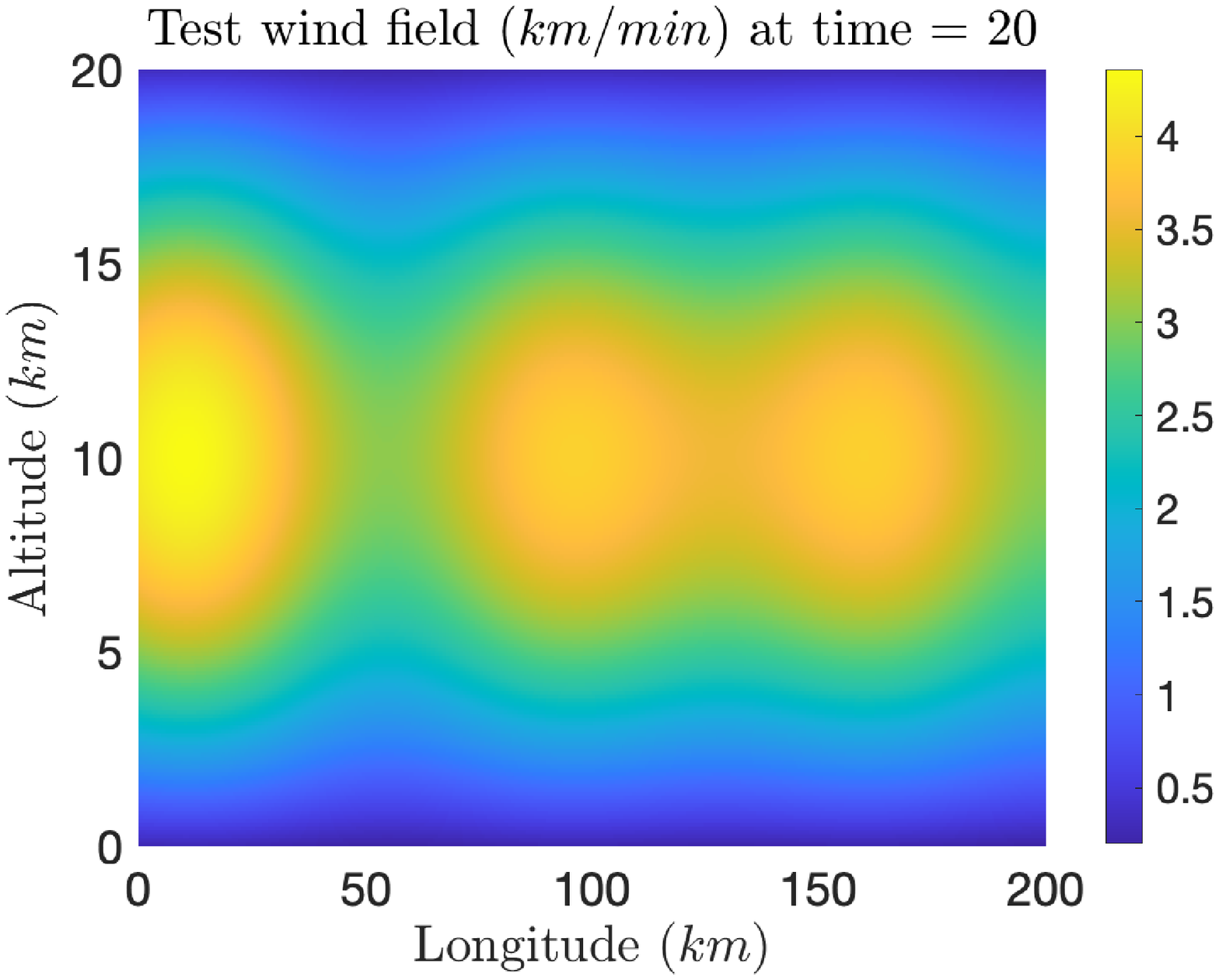}
    \includegraphics[width=0.32\textwidth]{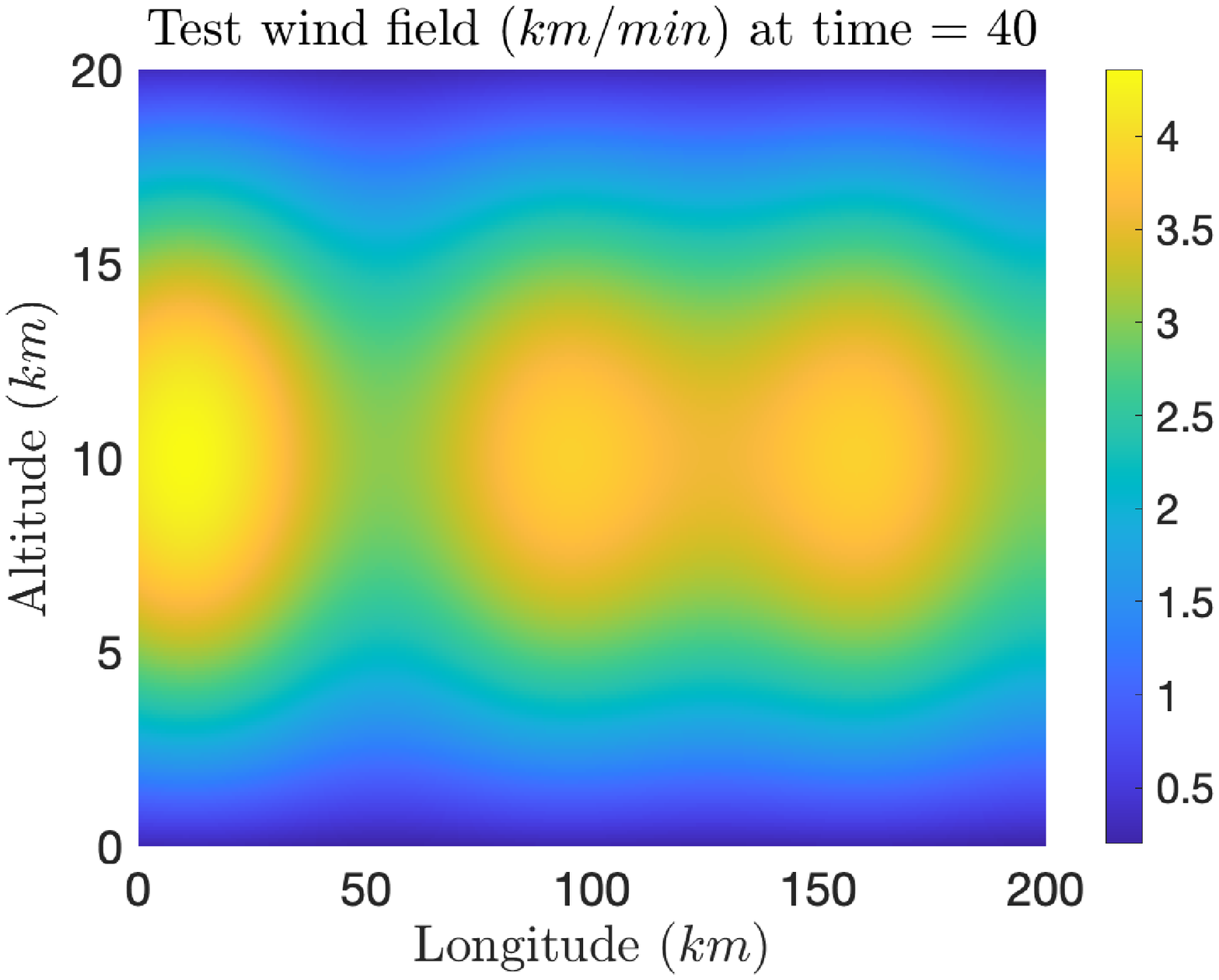}
    \includegraphics[width=0.32\textwidth]{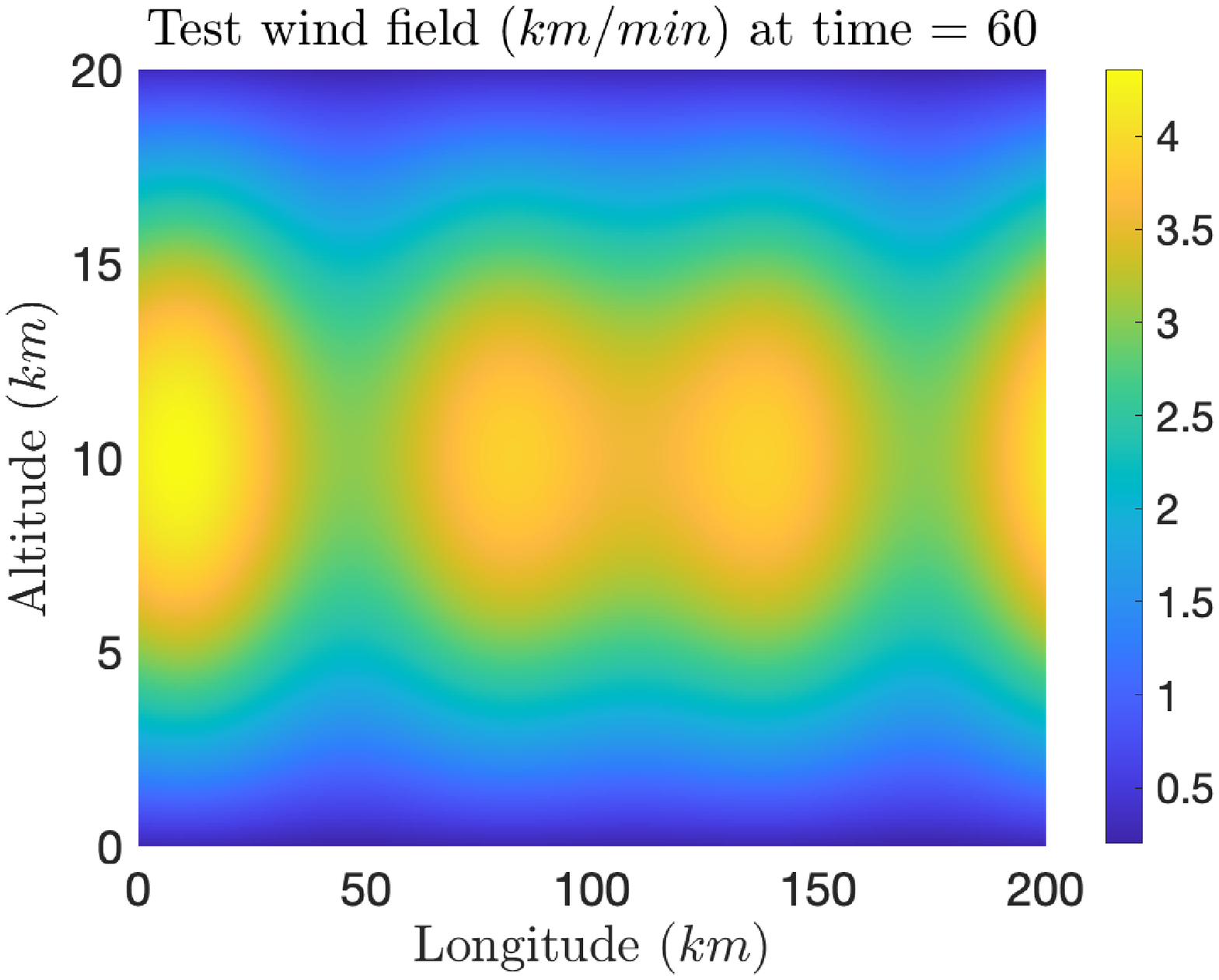} 
    \caption{Time snapshots of the mean (\emph{top}) and testing (\emph{bottom}) data wind fields $w$, which enter into the model through an advection field $\mathbf{v} = (w,0)$.} 
    \label{fig:wind_data}
\end{figure}

\begin{figure}
    \centering
    \includegraphics[width=0.4\textwidth]{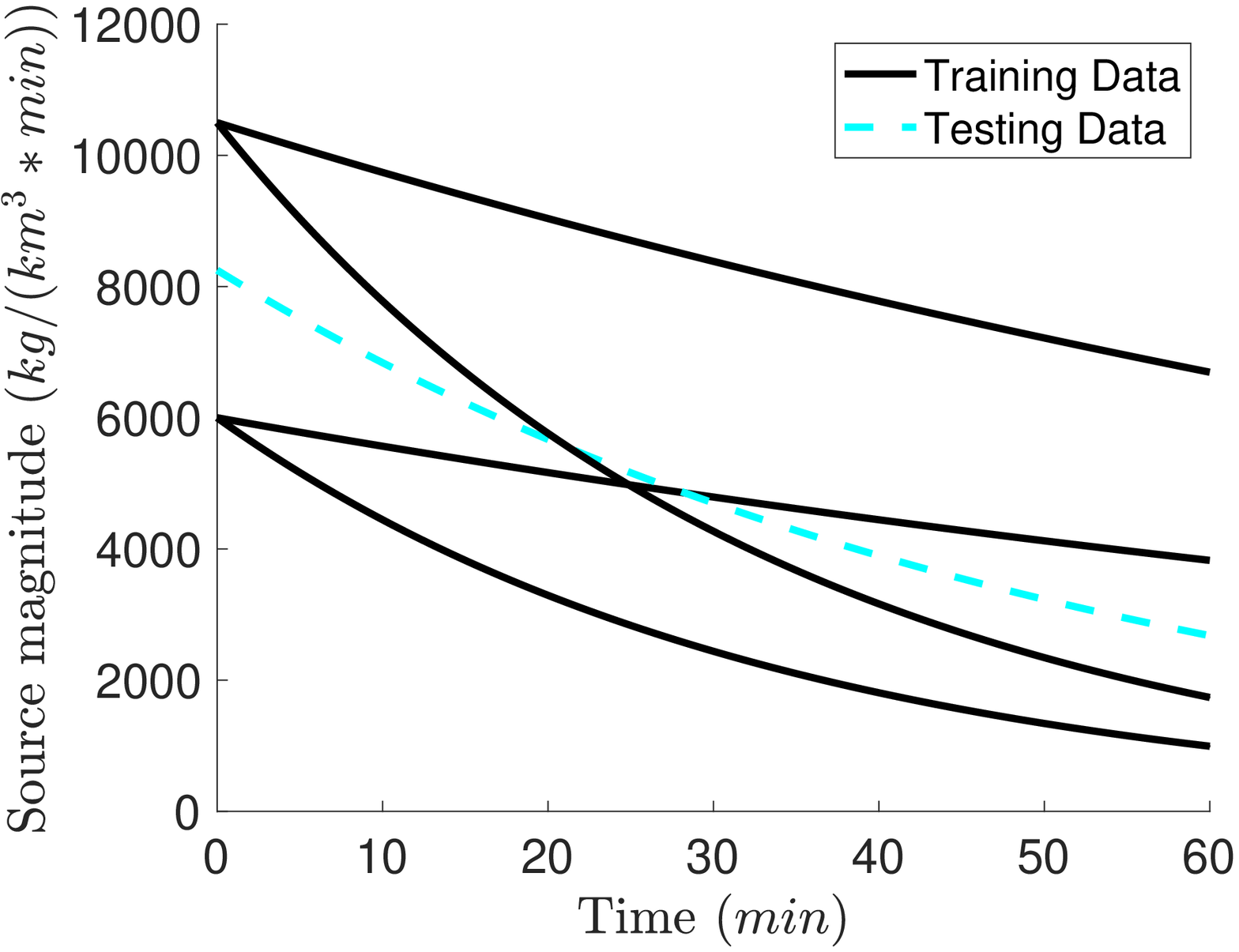}
    \includegraphics[width=0.4\textwidth]{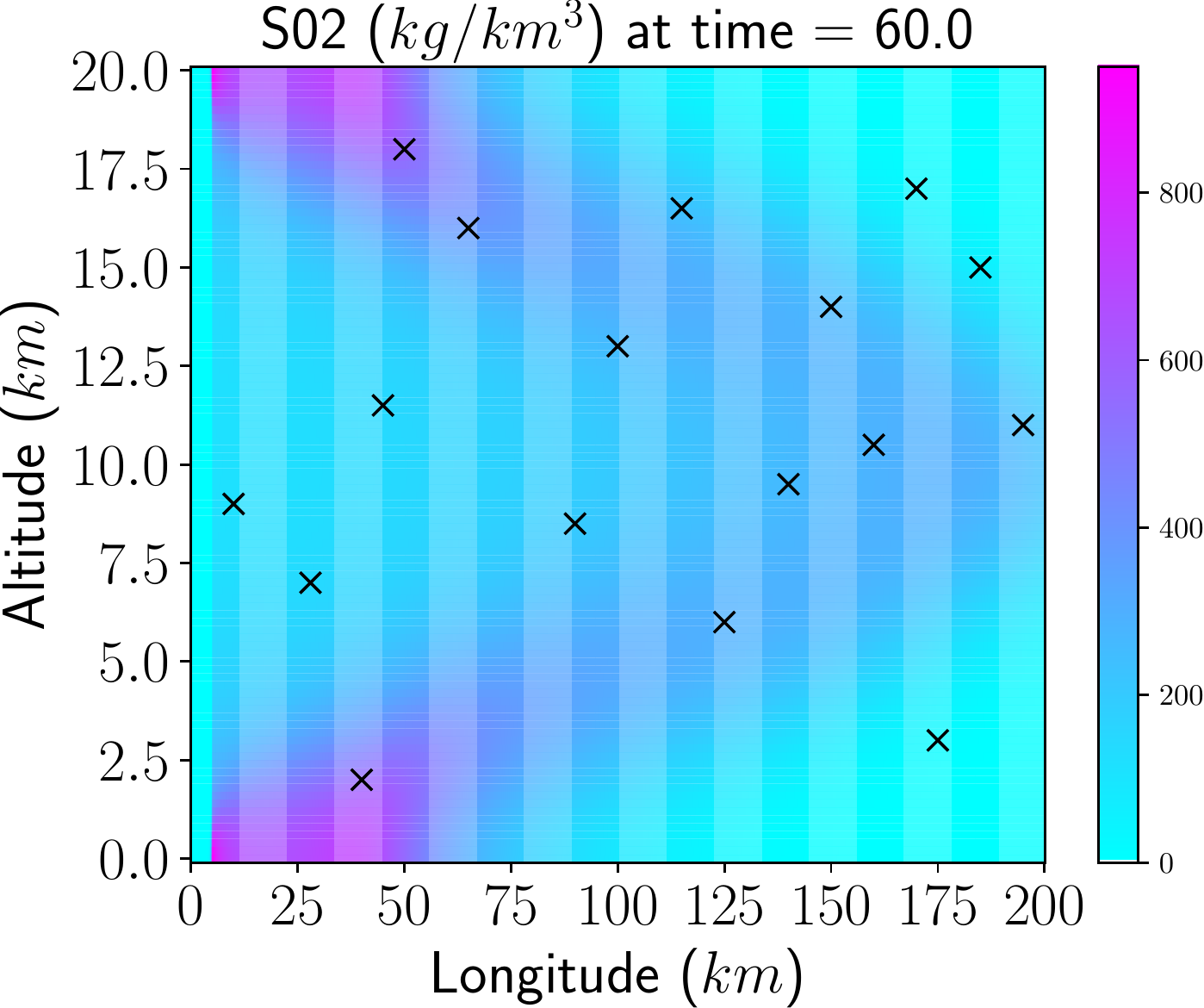}
    \caption{\emph{Left:} training and testing data for the source magnitude. \emph{Right:} testing data state solution with measurement locations marked by \SmallCross's.} 
    \label{fig:source_data}
\end{figure}

To demonstrate our inversion approach, we generated a test dataset which was never utilized during training or architecture tuning. For this test set, we choose a source magnitude which is different than those in the flow map training set, but is ``in the middle" of the training data, as shown in the left panel of Figure~\ref{fig:source_data}. To mimic realistic uncertainty in the wind field, we generate the test set wind field $\w^\star$ by generating many wind field samples from the same distribution as the training data and choosing the sample which maximizes the average distance (measured in the $\ell_2$ norm) from the training data wind fields. In other words, it is generated from the same distribution as the training data wind fields, but is chosen to ensure that it is ``not close to" any of them. Specifically, the relative $\ell^2$ distance between the test and mean wind field is $3.6\%$ and the relative $\ell^2$ distances between the test and training wind fields ranges from $4.7\%$ to $6.5\%$. The bottom row of Figure~\ref{fig:wind_data} displays time snapshots of the test data wind fields. After solving~\eqref{eqn:adv_diff_reaction_pde} with the testing data source and wind field, we contaminate the state data with multiplicative Gaussian noise whose mean is 1 and standard deviation is $0.02$, i.e. $2\%$ of the data magnitude. For the inversion, we only assume access to the state data at sparse spatial locations (but all time steps) as depicted in the right panel of Figure~\ref{fig:source_data}. Our inversion does not use any explicit knowledge of the source or wind fields used to generate the test dataset.

Using the sparse and noisy observations of the state (SO2 concentration), we seek to infer the source magnitude. To mimic realistic uncertainties, we use the mean wind field $\overline{\w}$, shown in the top row of Figure~\ref{fig:wind_data}, as our best estimate of the wind despite the data being generated by the testing wind field $\w^\star$ shown in the bottom row of Figure~\ref{fig:wind_data}. This erroneous wind field specification is characteristic of challenges faced in practice. As we show below, failing to account for this uncertainty yields poor inversion results; however, using the BAE approach allows for reliable inversion even when the wind field used in the inverse problem does not correspond to the wind field which generated the test data. 

\subsubsection*{Bayesian modeling}
\begin{figure}
    \centering
    \includegraphics[width=0.4\textwidth]{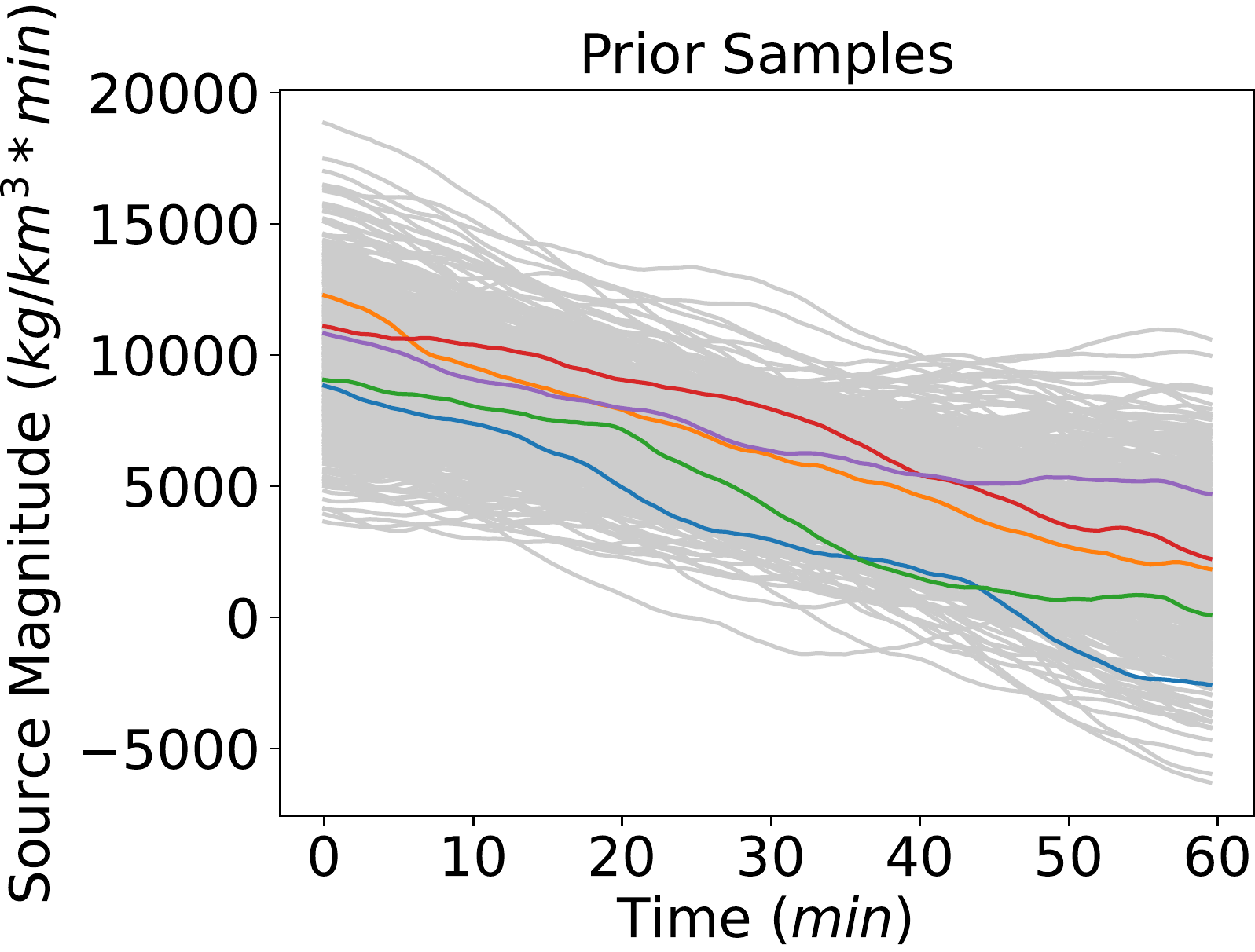}
    \caption{Prior samples of the source magnitude $z(t)$. Five samples are colored to aid in visualizing their smoothness while the others are grey scale to demonstrate their range of magnitudes.} 
    \label{fig:prior}
\end{figure}

We adopt the common assumption of additive Gaussian noise. The test data was contaminated by multiplicative Gaussian noise, but we do not assume knowledge of this in our formulation of the inverse problem. We define the noise covariance matrix as $\mathbf{\Gamma}_{\text{noise}} = (5^2) \mathbf{I}$, where $\mathbf{I}$ denotes the identity matrix. The noise standard deviation of $5$ is chosen to be approximately $4\%$ of the average state data magnitude, where $4\%$ is determined by adding $2\%$ noise and $2\%$ error in the flow map approximation. We define the prior on $\mathbf{z}$ with mean $\mathbf{\mu}=10,000-\frac{8,000}{60}\mathbf{t}$, where $\mathbf{t}$ is the vector of time nodes in the discretization. The prior covariance is defined as the squared inverse of a Laplacian-like differential operator~\citep{stuart_inv_prob}. Samples from the prior are shown in Figure~\ref{fig:prior}. We note that this prior is informed in that we assume knowledge of the decreasing magnitude, which is characteristic in the training and test data. However, we do not assume knowledge of the parametric form of how the test data was generated. As seen in Figure~\ref{fig:prior}, The prior covariance is defined using large covariance to emphasize a lack of prior knowledge on the source magnitude. 

\subsubsection*{Results}
There is modeling error due to our misspecification of the wind field, i.e. fixing the wind field to its mean $\overline{\w}$ rather than the actual wind $\w^\star$ which generated the testing data. To illustrate the importance of the BAE framework, we solve the inverse problem with both the BAE formulation and using the traditional formulation. In the BAE scenario, we use $b=50$ samples to estimate the mean and covariance of the model error distribution corresponding to fixing the wind field to its mean $\overline{\w}$.

\begin{figure}
    \centering
    \includegraphics[width=0.4\textwidth]{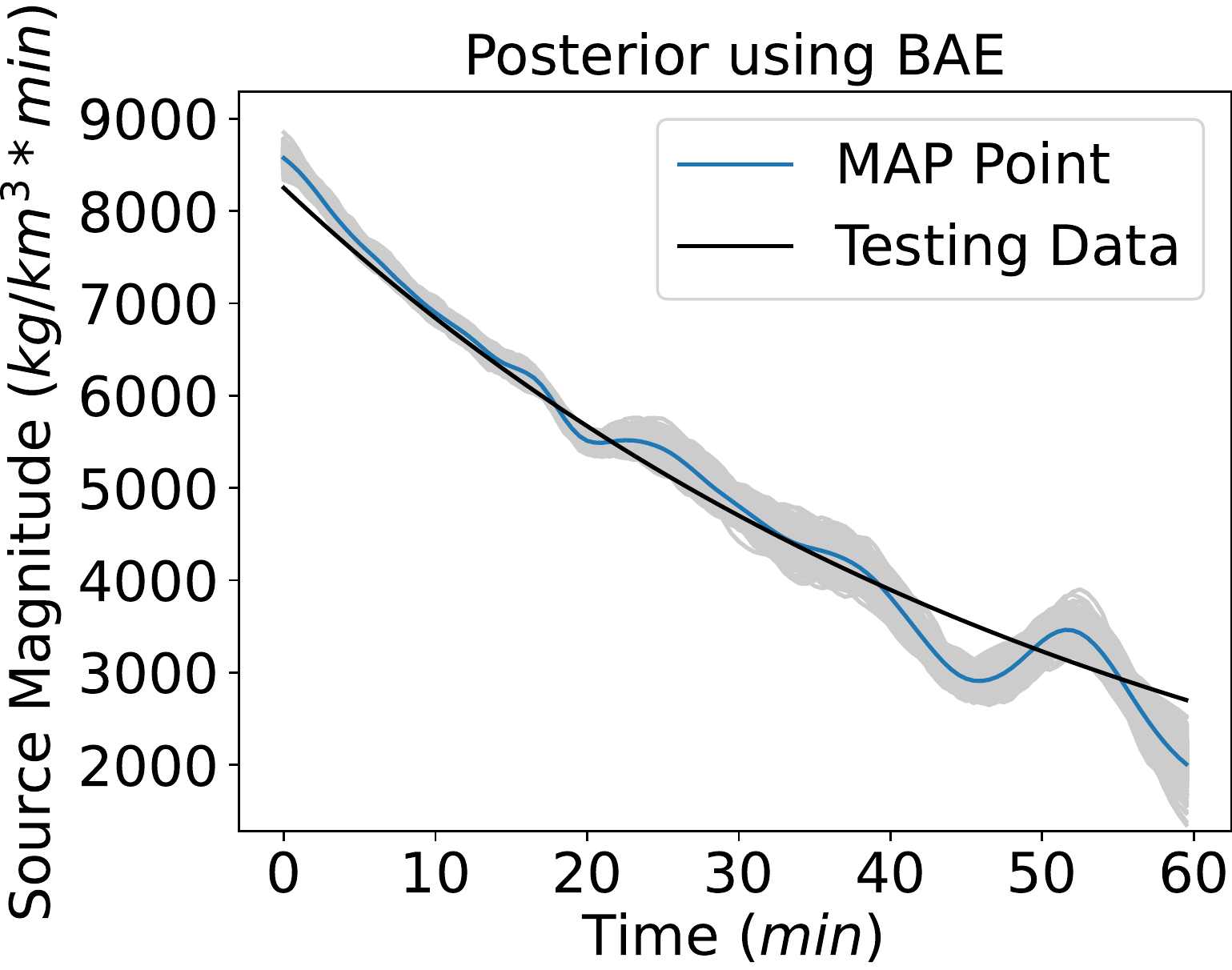} \hspace{5 mm}
    \includegraphics[width=0.4\textwidth]{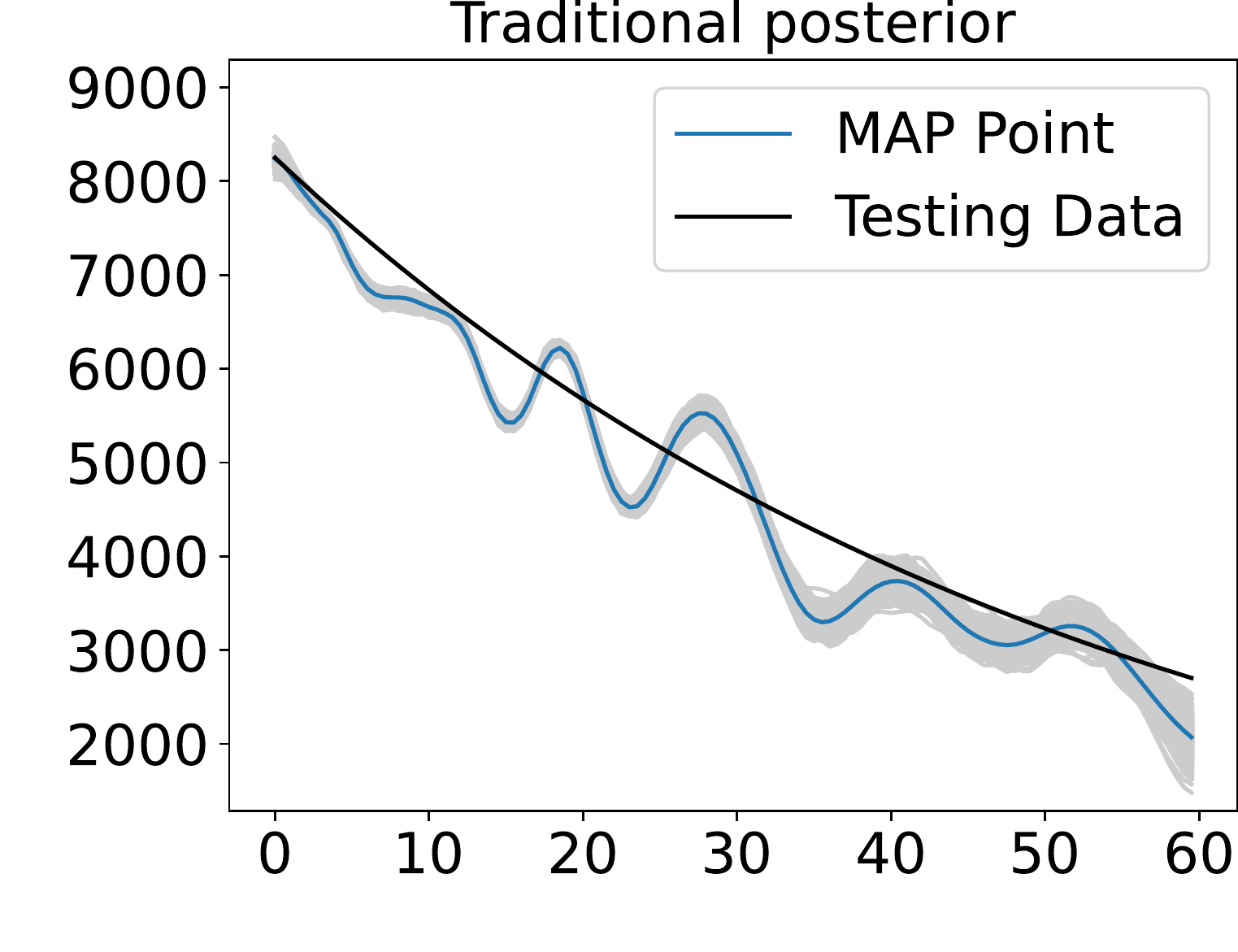}
    \caption{MAP point and posterior samples in the \textbf{lesser wind variability} case with the testing data source overlaid for comparison. \emph{Left:} solution using BAE. \emph{Right:} traditional posterior.} 
    \label{fig:posterior}
\end{figure}

The MAP point and posterior samples are shown in Figure~\ref{fig:posterior} with the left and right panels displaying the solution with and without using BAE, respectively. We observe that the posterior in the traditional Bayesian approach has oscillations due to modeling error which is not characteristic of the testing data, and that it has a small variance about this solution indicating confidence in an erroneous MAP point. This result is unsurprising (given that the wind field was misspecified in the model) and highlights the importance of BAE. Since our flow map approximation captured characteristics of how wind variability induced state variability, the likelihood update scaled the noise covariance so that the BAE formulation of the inverse problem avoided the oscillations and overconfidence seen in the traditional posterior. This highlights the utility of incorporating the wind into our flow map training so that it could be properly handed in the inversion. 

\subsubsection*{Increasing wind variability}
In a second study, the wind field variability (or range of uncertainty) is increased to understand limitations of the proposed approach. We repeat our previous data generation approach by executing $M=16$ PDE solves to generate a new training dataset. In this case, we use the same forcing magnitudes but sample new wind fields with greater variability by sampling $\theta$ from the uniform distributions specified in Table~\ref{tab:theta_parameters}. For these new wind fields, the relative $\ell_2$ differences between the training samples and mean wind range from $9.7\%$ to $10.8\%$ (recall a range of $3.3\%$ to $3.7\%$ in the lesser wind variability case).

Using the same flow map architecture from the previous study, we train a flow map approximation using this higher wind variability training set. With increased error in the flow map approximation, we define the noise covariance as $\mathbf{\Gamma}_{\text{noise}}=(8.5)^2 \mathbf{I}$ which has a larger standard deviation than before due to larger errors in the flow map approximation. We construct a new test dataset in the same way as before, with the exception that the test wind field is sampled with greater variability. The relative $\ell^2$ distance between the mean wind field and this new test wind field is $10.5\%$ and the relative $\ell^2$ distance between the test and training wind fields range from $15.6\%$ to $18.5\%$.

\begin{figure}
    \centering
    \includegraphics[width=0.4\textwidth]{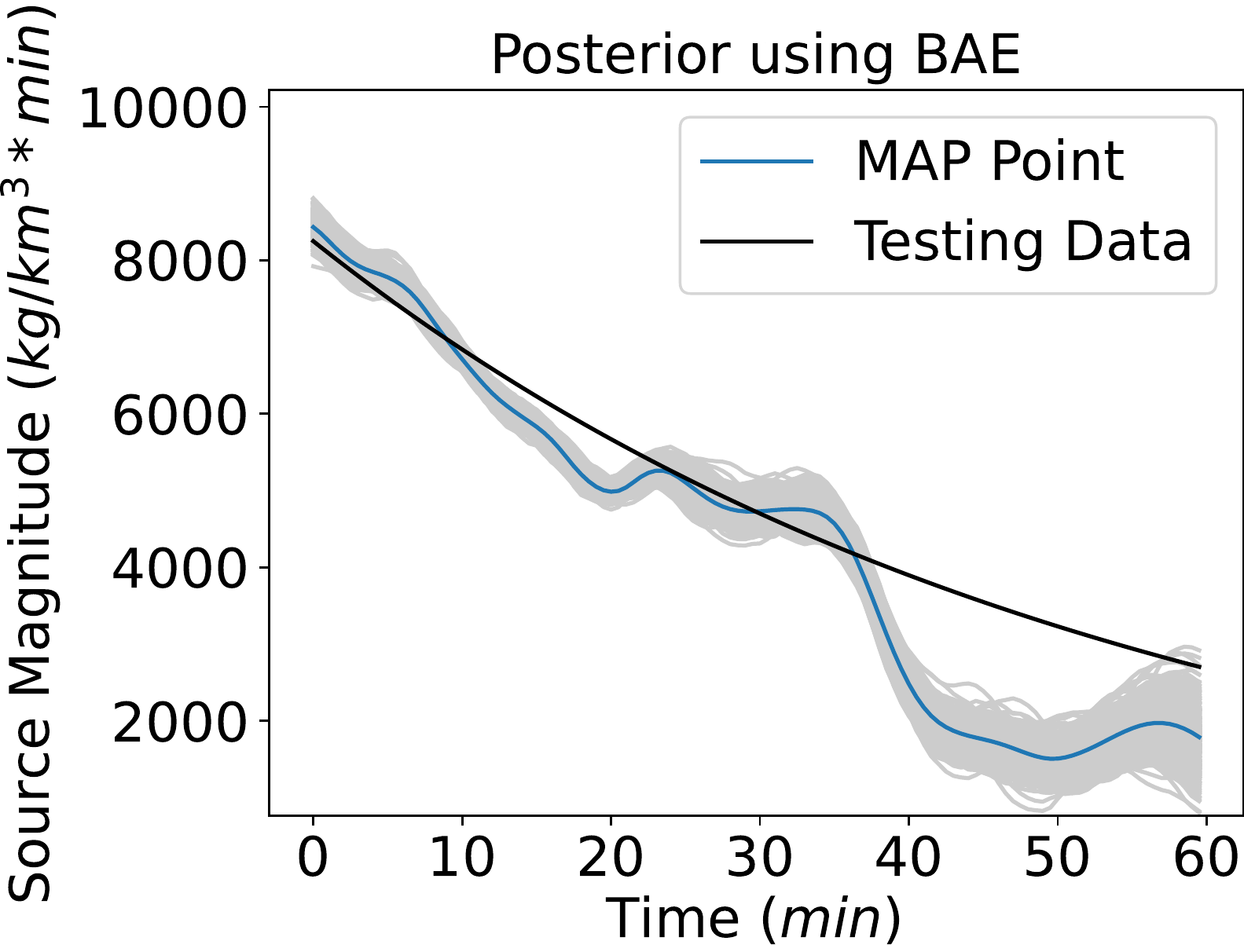} \hspace{5 mm}
    \includegraphics[width=0.4\textwidth]{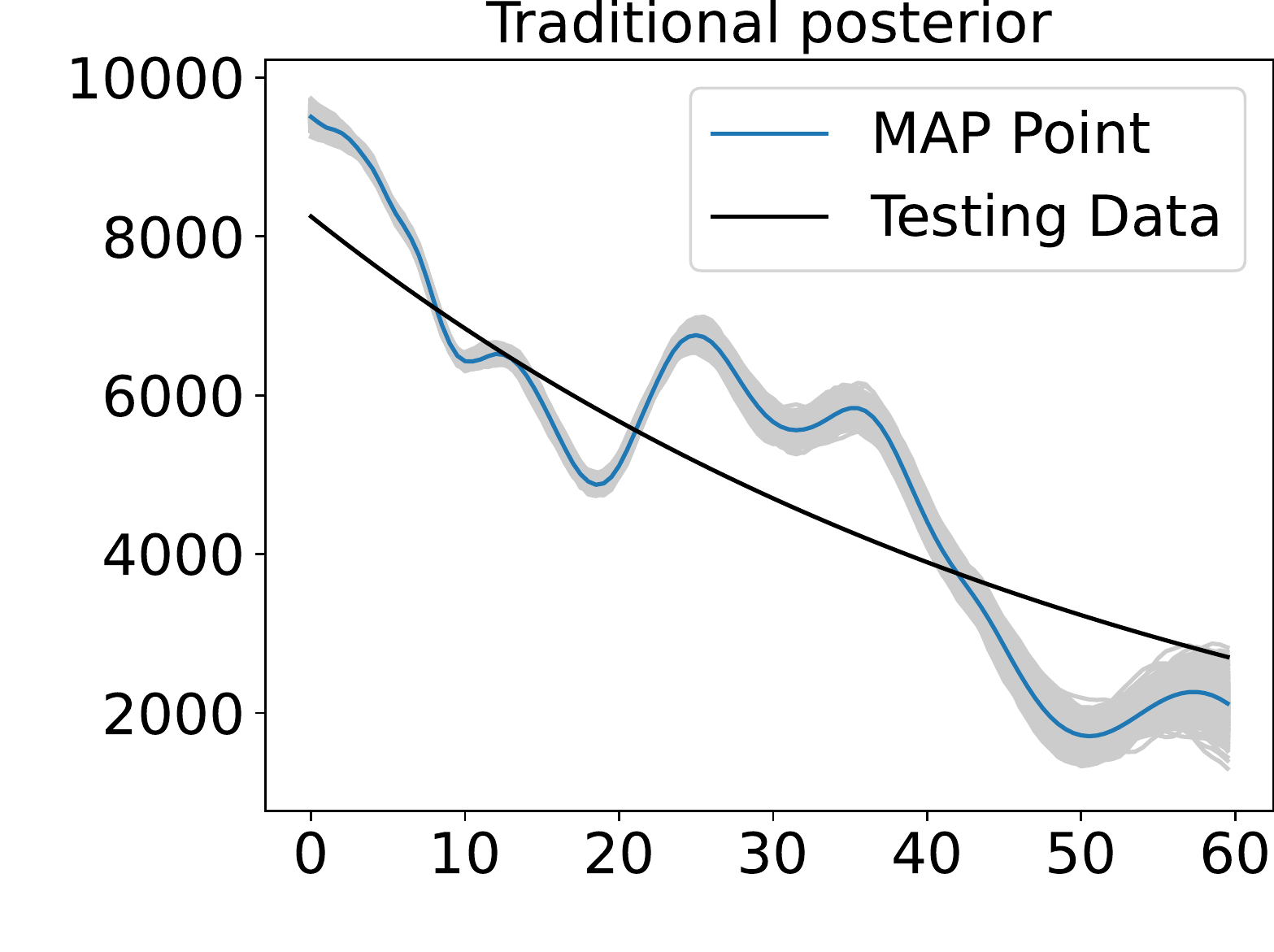}
    \caption{MAP point and posterior samples in the \textbf{greater wind variability} case with the testing data source overlaid for comparison. \emph{Left:} solution using BAE. \emph{Right:} traditional posterior.} 
    \label{fig:posterior_10pct}
\end{figure}

Figure~\ref{fig:posterior_10pct} compares the posterior using BAE and the traditional Bayesian formulation for this increased wind variability case. With increased wind variability, we observe greater error in the flow map approximation and inversion, as is expected. Nonetheless, we draw similar conclusions as before which again highlight the benefit of BAE. Comparing the inversion between the lesser and greater wind variability cases, we see the effect of wind variability on the inversion, but emphasize the quality of the result as the posterior shown in Figure~\ref{fig:posterior_10pct} is based on using only $M=16$ PDE solves to generate training data for this more difficult scenario. 

\begin{figure}
    \centering
    \includegraphics[width=0.4\textwidth]{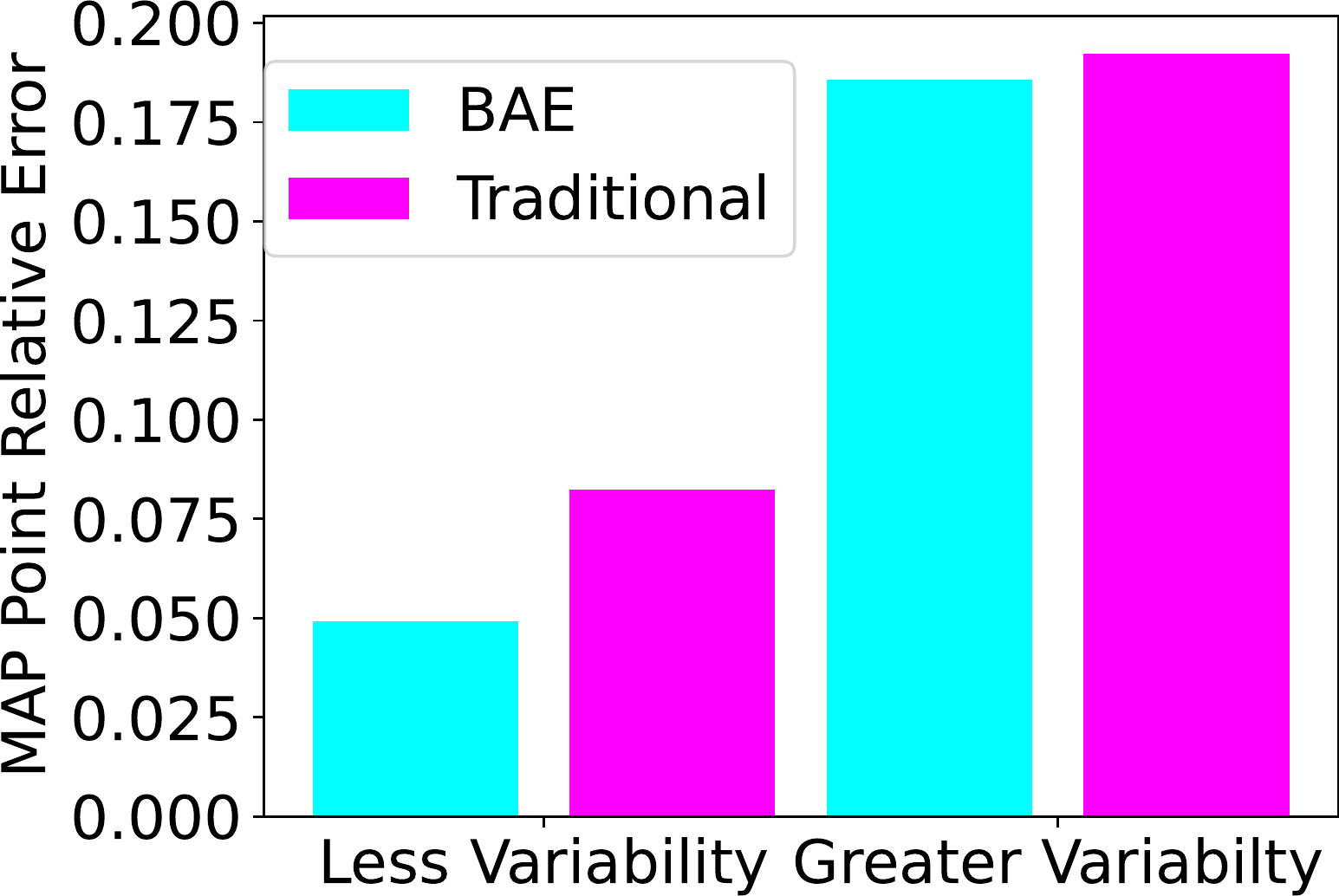} \hspace{5 mm}
    \includegraphics[width=0.4\textwidth]{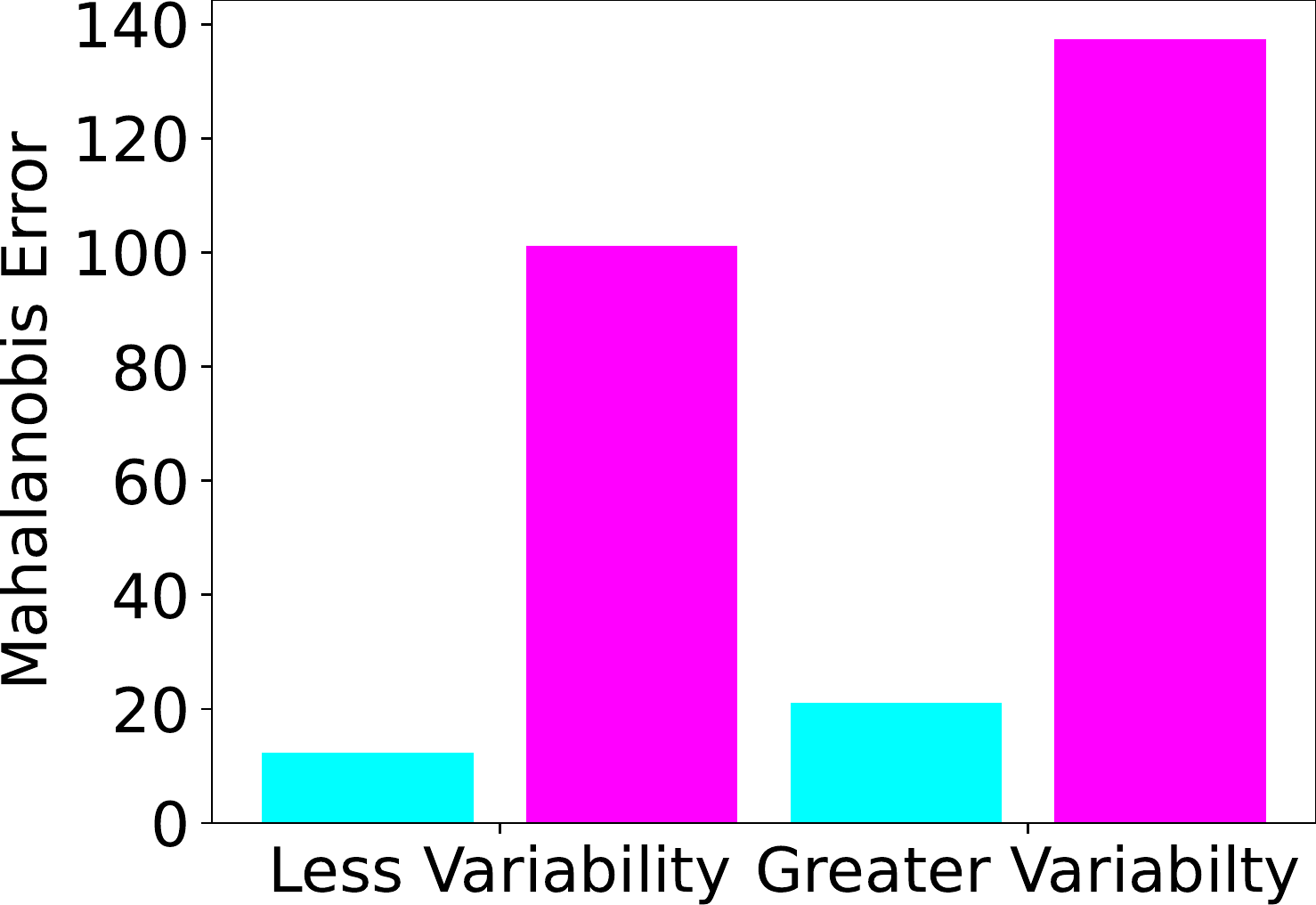} 
    \caption{Bar plots comparing the error estimating the test data source. \emph{Left panel:} the relative $\ell^2$ error between the MAP point and the test data source. \emph{Right panel:} the Mahalanobis distance between the posterior and the test data source. In each panel, the two leftmost (rightmost) bars correspond to the error on the lesser (greater) wind variability case with the bar color distinguishing the solution using BAE and the traditional Bayesian formulation.} 
    \label{fig:posterior_error_comparison}
\end{figure}

To quantify the quality of the posterior distribution, we compute metrics on the solution for the four cases, lesser and greater wind variability using BAE and using the traditional Bayesian formulation. Two metrics are considered. The first is the relative $\ell^2$ error between the MAP point and the test data source. The second is the Mahalanobis distance between the posterior distribution and the test data source. This is a commonly used metric to measure distance between a probability distribution and a point. It can be viewed as a vector-valued generalization of measuring how many standard deviations a point is away from the distribution's mean, i.e., the standard score or $z$-score of the test point with respect to the posterior distribution \citep{maupin2018validation}. As such, it is the relevant metric for evaluating the quality of the posterior with respect to the ground truth. For a posterior distribution $\pi_{\text{post}}$ with mean $\mathbf{\mu} \in \R^N$ and covariance $\mathbf{\Sigma} \in \R^{N \times N}$, the Mahalanobis distance from $\pi_{\text{post}}$ to a point $\mathbf{x} \in \R^N$ is defined as
\begin{align}\label{eq:mahalanobis_distance}
d(\mathbf{x}, \pi_{\text{post}})=
\sqrt{ \left( \mathbf{\mu} - \mathbf{x} \right)^\top \mathbf{\Sigma}^{-1} \left( \mathbf{\mu} - \mathbf{x} \right) }.
\end{align}
The left panel of Figure~\ref{fig:posterior_error_comparison} shows the relative $\ell^2$ error between the MAP point and the test data source.
The right panel of Figure~\ref{fig:posterior_error_comparison} shows the Mahalanobis distance between the Laplace approximation of the posterior and the test data source. In both panels, we present results for the lesser wind variability and the greater variability datasets.
For both the $\ell^2$ error metric and the Mahalanobis distance, there is an increase in error that comes from greater wind variability. 
In the case of $\ell^2$ error, we see only a marginal improvement using the BAE posterior compared to the traditional posterior. 
This belies the significant advantage of the BAE approach that is visible in Figures \ref{fig:posterior} and \ref{fig:posterior_10pct}.
In comparing the two approaches using the Mahalanobis distance in the right panel, we observe quantification of the idea that the traditional Bayesian posterior was overconfident in the wrong solution.
The much larger Mahalanobis distance for the traditional posterior demonstrates the significant advantage of BAE to provide uncertainty quantification, which is crucial in limited data settings.

\subsubsection*{Presence of multiple MAP points}
\begin{figure}
    \centering
    \includegraphics[width=0.4\textwidth]{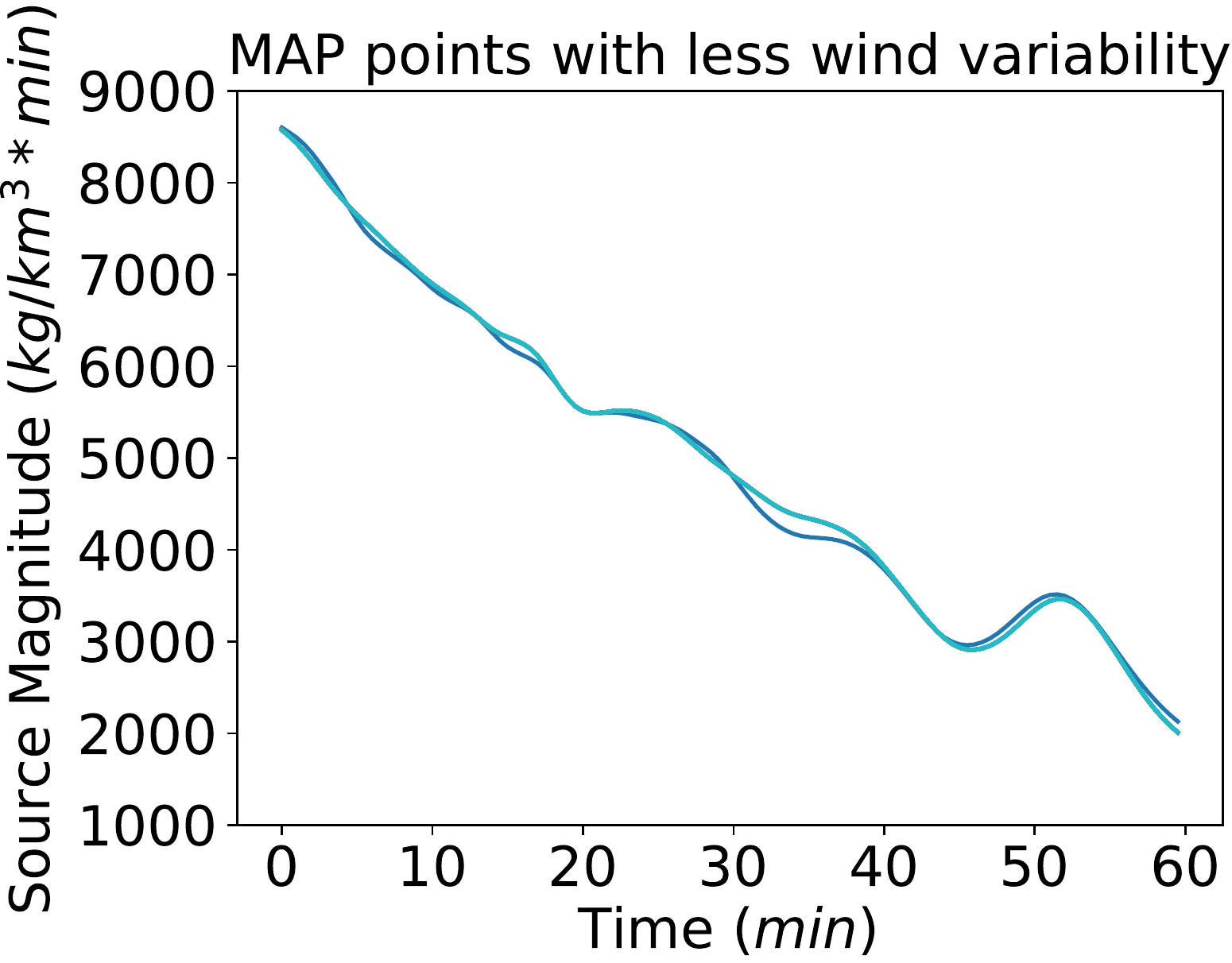} \hspace{5 mm}
    \includegraphics[width=0.41\textwidth]{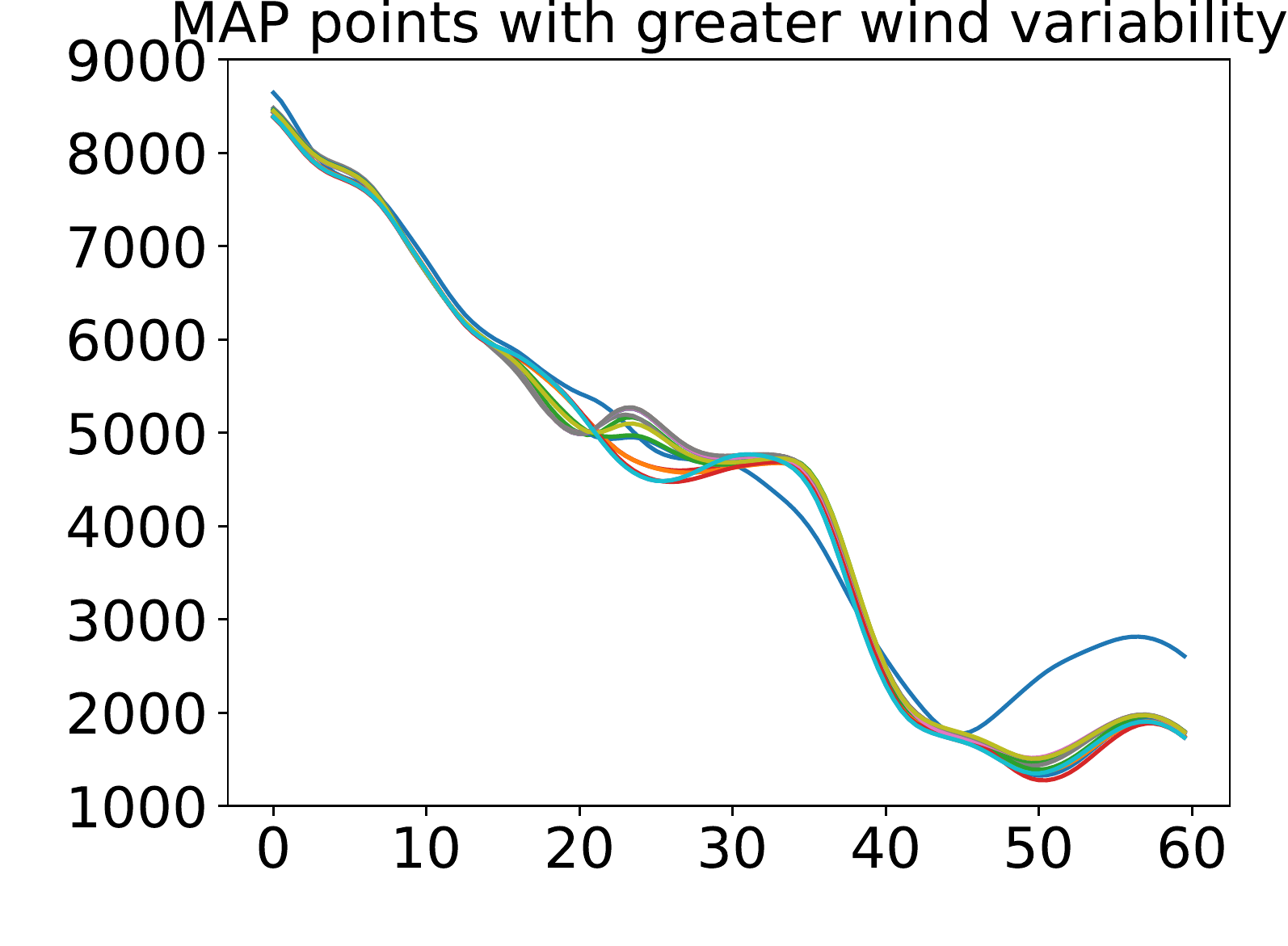} 
    \caption{MAP points determined by 20 different optimization executions starting from random perturbations of the initial condition. The left (right) panel displays the MAP points in the less (greater) wind variability case.} 
    \label{fig:map_variability}
\end{figure}

If the true flow map of the PDE model \eqref{eqn:adv_diff_reaction_pde}, rather than an approximation, were used to define the parameter-to-observable map, the MAP point would be unique (since the source-to-state solution mapping is linear for a fixed wind field). However, the DNN flow map approximation is nonlinear. While this is appropriate to capture the nonlinear effect of the wind and for wider applicability to nonlinear problems, a consequence is that there may be multiple MAP points corresponding to different local minima found in by the optimization algorithm. The results presented so far have all been for single MAP points. To understand the stability of the MAP point approximation, we restarted the MAP point optimization algorithm from distinct, randomly perturbed initial iterates. Figure~\ref{fig:map_variability} shows the 20 MAP points (some overlaying one another) in the BAE formulation for both the lesser and greater wind variability cases. We observe that in the lesser wind variability case, we found two MAP points, i.e. two local minima determined by the 20 optimization restarts, which are close to one another. However, due to the increased complexity of the greater wind variability case, there are a greater number of MAP points with more significant differences between them. The existence of several MAP points, and the effect on source inversion, should be kept in mind even for linear problems using the proposed approach. 

\section{Conclusion} \label{sec:conclusion}
We have presented a pragmatic approach which combines learning data-driven surrogate operators parametrized by DNNs with Bayesian inversion and Bayesian approximation error to estimate uncertain parameters/processes in geophysical systems. We leveraged algorithmic differentiation to efficiently compute gradients and Hessian approximations which enabled optimization and approximate posterior sampling without the need for extensive algorithmic parameter tuning to achieve convergence. The combination of these tools enabled rapid source inversion with uncertainty quantification in a novel setting where inversion is confounded by auxiliary unobserved processes, high state dimension, and small ensembles of training data, while not requiring intrusion to obtain derivative information. The methodology is thus transferable to different applications and allows for assimilation of stored runs from databases of historical simulations. 

This represents an important step toward enabling inversion for more complex models with an eye toward applying the tools to earth system models where the data may be very limited due to computational complexity. Throughout our developments, choices were guided by the requirement of scalability to global models. Our results presented a simplified regional aerosol transport model to enable thorough numerical exploration. To find success at the scale of earth system models, which contain many state variables, it will be critical to learn spatiotemporal dynamics through subsets of the state space. Such subsets of the state space is associated with the predominate physical processes which influence the parameter of interest. Their determination is a topic of ongoing research.

Using DNN operator approximations for large-scale spatiotemporal processes provide many opportunities but also their own challenges. Data processing considerations such as the choice of time resolution and spatial dimension reduction approach require more complete analysis. Ongoing work is exploring the use of tensor decompositions to incorporate spatiotemporal structure more naturally in our approach. As highlighted in our numerical results, DNNs introduce additional variability both in the training process (getting different networks by retraining on the same data) and in the inversion process (the presence of additional MAP points as a result of network approximation error). Understanding the limitations and failure modes of the approach remains critical to ensure reliable conclusions. However, their prospect as a pragmatic approach makes neural network operator approximations appealing as a topic of ongoing research.

\section{Acknowledgements}

This article has been authored by an employee of National Technology \& Engineering Solutions of Sandia, LLC under Contract No. DE-NA0003525 with the U.S. Department of Energy (DOE). The employee owns all right, title and interest in and to the article and is solely responsible for its contents. The United States Government retains and the publisher, by accepting the article for publication, acknowledges that the United States Government retains a non-exclusive, paid-up, irrevocable, world-wide license to publish or reproduce the published form of this article or allow others to do so, for United States Government purposes. The DOE will provide public access to these results of federally sponsored research in accordance with the DOE Public Access Plan\footnote{\texttt{https://www.energy.gov/downloads/doe-public-access-plan}}. This work was supported by the Laboratory Directed Research and Development program at Sandia National Laboratories, a multimission laboratory managed and operated by National Technology and Engineering Solutions of Sandia LLC, a wholly owned subsidiary of Honeywell International Inc. for the U.S. Department of Energy’s National Nuclear Security Administration under contract DE-NA0003525. SAND2023-00818O. 

\appendix

\section{$\text{SO}_2$  Data Generation} \label{appendix:sim_model}
The $\text{SO}_2$ dispersion data is generated by solving the PDE~\eqref{eqn:adv_diff_reaction_pde}
where $\kappa$ is the diffusion coefficient, $\mathbf{v}= ( w(x,y,t),0 )$ is the wind field,
\begin{align*}
S \mathbf{e}_y = (0,S) = \left( 0,\sqrt{ \frac{8}{3} \frac{\rho_{\text{SO}_2}}{\rho_{\text{atmo}}} \frac{g}{C_s} r } \right)
\end{align*}
is the effect of the particles falling due to gravity with a terminal speed $S$ (which is calculated by setting the force of gravity equal to the force of drag), and $\mathcal R(u) = -ku$ is the reaction function modeling chemistry (with $k$ being the e-folding time).

The longitudinal wind field is given by
\begin{align*}
 w(x,y,t) = X(x,t) Y(y,t)
 \end{align*}
where
\begin{multline}
X(x,t) = 1 + \theta_6 \sin \left( \frac{4 \pi T_1(t) x}{200} \right) + \theta_7 \cos \left( \frac{4 \pi T_1(t) x}{200} \right) \\
+ \theta_8 \sin \left( \frac{6 \pi T_2(t) x}{200} \right) + \theta_9 \cos \left( \frac{6 \pi T_2(t) x}{200} \right) 
\end{multline}
with
\begin{align*}
T_1(t) = 0.9 + \theta_1 \cos \left( \frac{2 \pi \theta_2 t}{60} \right), \qquad T_2(t) = 0.9 + \theta_3 \cos \left( \frac{4 \pi \theta_4 t}{60} \right), 
\end{align*}
and
\begin{align*}
Y(y,t) =  0.25 + 3.75 T_3(t) \sin \left( \frac{\pi y}{20} \right),
\end{align*}
with
\begin{align*}
T_3(t) = 0.9 + 0.1 \theta_5 \cos \left( \frac{2 \pi t}{60} \right).
\end{align*}
The physical model parameters are given in Table~\ref{tab:parameters} and the range of values from which we sample the hyperparameters $\mathbf{\theta}=(\theta_1,\theta_2,\dots,\theta_9) \in \R^9$ defining the wind field is given in Table~\ref{tab:theta_parameters}.

\begin{table}[!ht]
\centering
\begin{tabular}{c|c|c|c}
\hline 
$\kappa$ & Diffusivity of $\text{SO}_2$ & $\text{km}^2/\text{min}$ & $1 \times 10^{-1}$ \\
$w(x,y,t)$ & Longitudinal wind velocity & $\text{km}/\text{min}$ & See equation \\
$S$ & Terminal falling speed & $\text{km}/\text{min}$ & $1.705 \times 10^{-4}$ \\
$\rho_{\text{SO}_2}$ & Density of $\text{SO}_2$ & $\text{kg}/\text{m}^3$ & 2.92 \\
$\rho_{\text{atmo}}$ & Density of the atmosphere & $\text{kg}/\text{m}^3$ & 1.28 \\
$g$ & Acceleration due to gravity & $\text{km}/\text{min}^2$ & 35.316 \\
$C_s$ & Drag coefficient & $-$ & $.38$ \\
$r$ & Radius of $\text{SO}_2$ particle & $\text{km}$ & $5 \times 10^{-11}$ \\
$\mathcal R(u)$ & Reaction function for $\text{SO}_2$ chemistry & $\text{kg}/(\text{km}^3 * \text{min})$ & See equation \\
$k$ & e-folding time for $\text{SO}_2$ depletion & $\text{min}^{-1}$ & $1/44,640$ \\
$f$ & Source term modeling $\text{SO}_2$ injection & $\text{kg}/(\text{km}^3 * \text{min})$ & See equation \\ 
\hline 
\end{tabular}
\caption{Model parameters.}
\label{tab:parameters}
\end{table}

\begin{table}[!ht]
\centering
\begin{tabular}{c|c|c|c|c|c|c|c|c|c}
Parameter & $\theta_1$ & $\theta_2$ & $\theta_3$ & $\theta_4$  & $\theta_5$  & $\theta_6$  & $\theta_7$  & $\theta_8$  & $\theta_9$   \\
\hline
Minimum & 0.0 & 0.95 & 0.0 & 0.95 & -0.1 & -0.05 & -0.05 &-0.05 &-0.05 \\
\hline
Maximum &0.2 & 1.05 & 0.2 & 1.05 & 0.1 & 0.15& 0.15& 0.15& 0.15  \\
\hline 
\end{tabular}
\caption{Ranges for $\theta$ samples corresponding to the greater wind variability case. In the lesser wind variability case, the interval for $\theta_k$, $k=1,2,\dots,9$, is scaled by $0.35$, for instance, the interval $[-0.1,0.1]$ is reduced to $[-0.035,0.035]$. We assume each $\theta_k$ is independent and uniformly distributed.}
\label{tab:theta_parameters}
\end{table}

\section{Flow Map Hyperparameter Tuning} \label{sec: appendix_flowmap}
\begin{figure}[h!]
    \centering
    \includegraphics[width=0.45\textwidth]{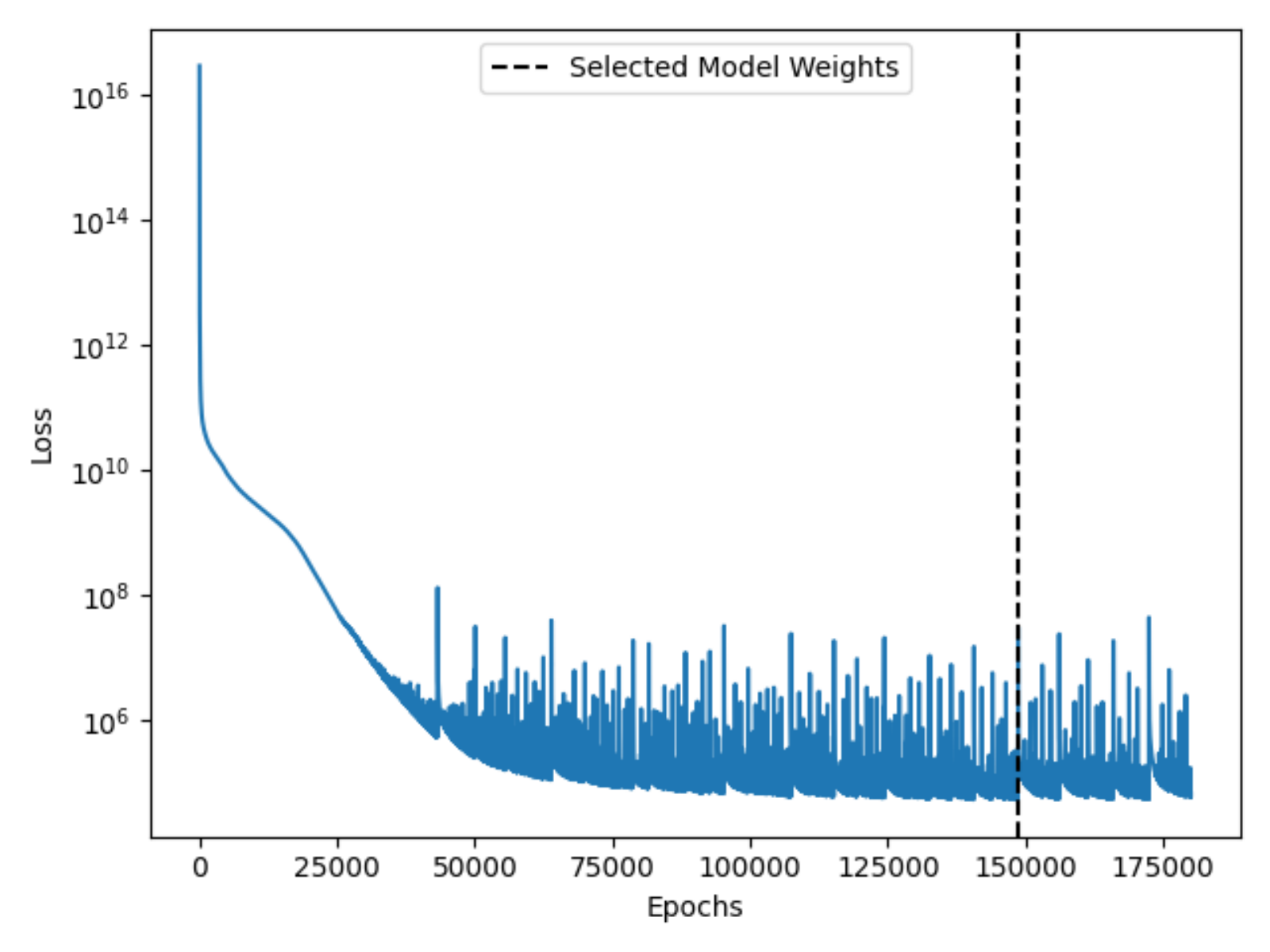}
    \includegraphics[width=0.45\textwidth]{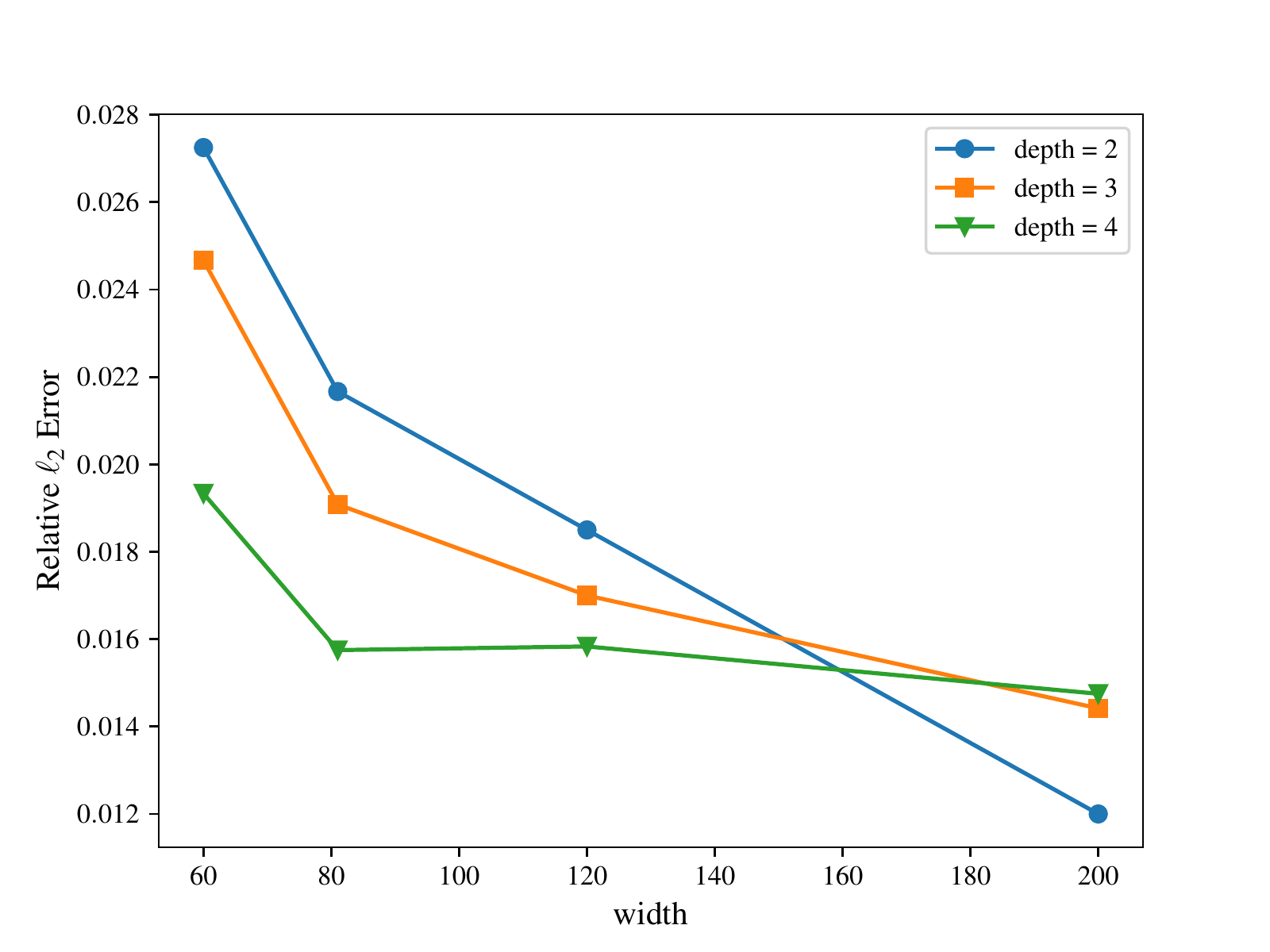}
    \caption{Left panel: example loss function over training epochs. Right panel: network performance on varying widths and depths. Results shown are averaged over 3 training instances with random Glorot network initializations.} 
    \label{fig:learning_loss}
\end{figure}

We detail the configurations used when training the flow map in Table~\ref{tab:hyp}.  An example of the learning curve during training is shown in the left panel of Figure~\ref{fig:learning_loss}. 

\begin{table}[h!]
\centering
\begin{center} 
\begin{tabular}{ c | c | c | c | c | c  }
   & {Tuning P, as in Fig.~\ref{fig:tune_P}}  & Tune Width \& Depth  & Final Flow Map\\
    \hline 
Epochs              & 180e3 & 180e3 & 180e3 \\
Learning Rate       & 0.002 & 0.0008 & 0.0008\\
Learning Decay Rate       & 0.04 & 0.04 & 0.04\\
Hidden Layer Width  & 81  & N/A & 200 \\
Hidden Layer Depth  & 2 & N/A & 2 \\
\end{tabular}
\end{center}
\caption{Hyperparameters used for all flow map computational studies. The flow map configuration after tuning that is used for the inversion problem is listed in the rightmost column.  }
\label{tab:hyp}
\end{table}

We also show the results of the width and depth tuning study in the right panel of Figure~\ref{fig:learning_loss}.  Networks with widths larger than 200 had minimal improvement in performance while increasing training time.

\bibliography{References}

\end{document}